\documentclass{article} 
\usepackage{iclr2027_conference,times}

\usepackage{amsmath,amsfonts,bm}

\def\eqref#1{equation~\ref{#1}}

\def\1{\bm{1}}

\DeclareMathAlphabet{\mathsfit}{\encodingdefault}{\sfdefault}{m}{sl}
\SetMathAlphabet{\mathsfit}{bold}{\encodingdefault}{\sfdefault}{bx}{n}

\usepackage{hyperref}
\usepackage{url}

\usepackage{booktabs}
\usepackage{multirow}
\usepackage{graphicx}
\usepackage{wrapfig}
\usepackage{float}
\usepackage{enumitem}
\usepackage{subcaption}
\usepackage{algorithm}
\usepackage{algpseudocode}
\title{Generalizable Spatio-Temporal Forecasting Beyond the Closed-World Setting}

\author{Shengnan Guo$^{1,2}$, Jinjun Cheng$^{1}$, Xiaowei Mao$^{1}$, Jianhao Zhang$^{1}$, Haochen Lv$^{1}$,\\
\bf Yang Jiang$^{1}$, YiHeng Huang$^{1}$, Yan Lin$^{4}$, Youfang Lin$^{1,3}$, Huaiyu Wan$^{1,3}$\thanks{Corresponding author.}\\
$^{1}$School of Computer Science and Technology, Beijing Jiaotong University, Beijing, China\\
$^{2}$Key Laboratory of Big Data \& Artificial Intelligence in Transportation,\\ Ministry of Education, Beijing, China\\
$^{3}$Beijing Key Laboratory of Traffic Data Mining and Embodied Intelligence, Beijing, China\\
$^{4}$Department of Computer Science, Aalborg University, Aalborg, Denmark\\
\texttt{\{guoshn, chengjinjun, maoxiaowei, jianhaozhang\}@bjtu.edu.cn}\\
\texttt{\{jiangyang, lvhaochen, huangyiheng, yflin, hywan\}@bjtu.edu.cn}\\
\texttt{lyan@cs.aau.dk}}

\iclrfinalcopy 
\begin{document}

\maketitle

\begin{abstract}
Spatio-temporal forecasting is fundamental to many real-world systems, yet deployment environments are inherently varying: temporal distributions may evolve, spatial domains may change, and observation availability may shift from training conditions. 
Existing models are often tightly coupled to their training environments through node-specific representations, predefined spatial support sets, and reliance on consistently available observations, limiting generalization to changing environments. 
To this end, we propose \textbf{VaryST}, a compact spatio-temporal forecasting model trained from scratch that achieves strong generalization by dynamically constructing predictive context from available long-term observations.
To accommodate temporal changes, VaryST derives node representations from long-term histories, capturing current behavior in the context of longer-term patterns.
To adapt to spatial changes, it dynamically selects functionally relevant support nodes beyond predefined neighborhoods and integrates their information with recent and long-term target states and global context.
To handle varying observation availability, VaryST selectively masks historical patches on which the model relies most during training, encouraging it to exploit alternative temporal and spatial dependencies for accurate forecasting when key observations become unavailable. 
Extensive experiments on six real-world datasets against seventeen baselines show that VaryST achieves strong in-domain forecasting accuracy, substantially improves zero- and few-shot generalization under spatial and temporal changes, and remains robust to varying observation availability.
\end{abstract}

\section{Introduction} 
Spatio-temporal (ST) forecasting predicts the future states of spatially distributed nodes by modeling temporal dynamics and spatial dependencies, supporting applications such as traffic management, energy systems, and environmental monitoring. Despite substantial progress, many existing methods operate within a largely \textit{closed-world setting}, assuming that temporal distributions, spatial domains, and observation availability remain stable between training and deployment.

\begin{figure}
    \includegraphics[width=\linewidth]{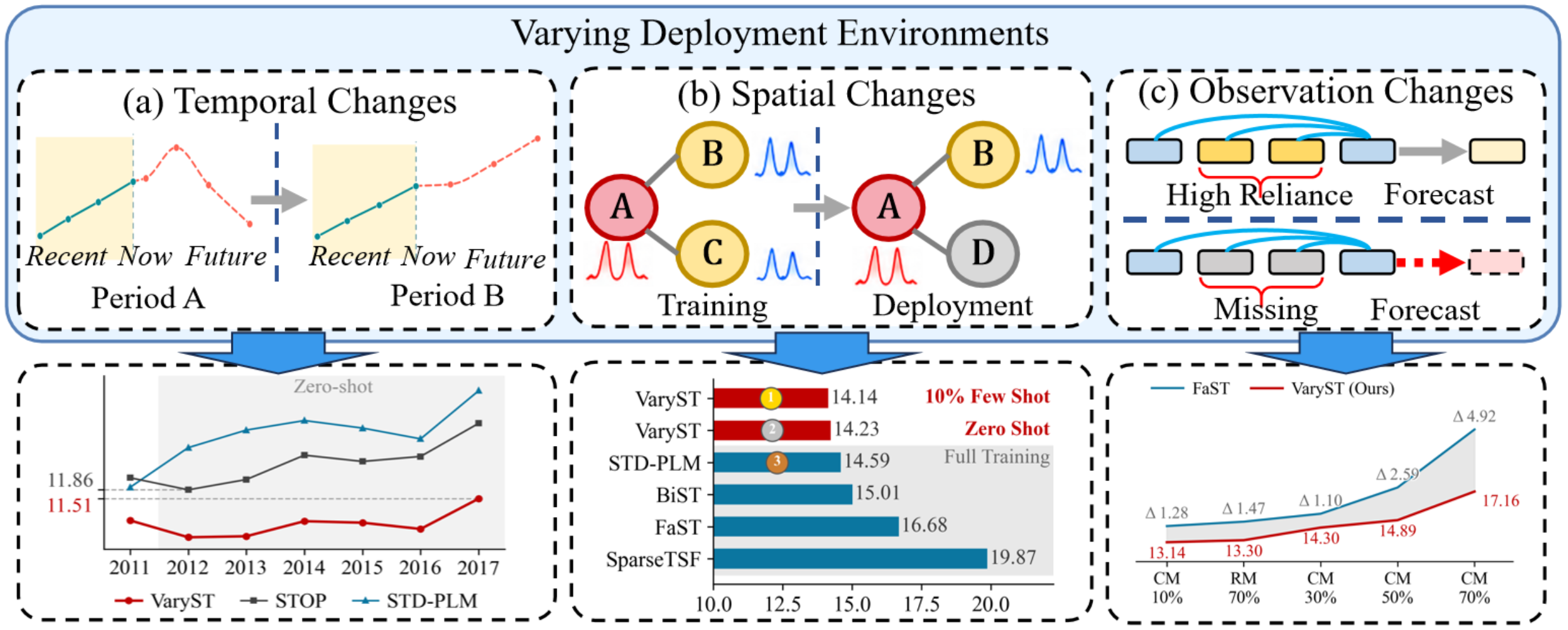}
    \captionsetup{skip=2pt}
    \caption{ST models in varying deployment environments}
    \label{fig:varst_intro}
\end{figure}

Real-world deployment, however, is inherently varying, as shown in Figure~\ref{fig:varst_intro}. {\textit{Temporal distributions}} evolve with seasonal variations and long-term trends. {\textit{Spatial domains}} change as models transfer to new cities or sensor networks expand. {\textit{Observation availability}} varies with sensor failures and communication interruptions, producing missing rates and patterns that differ from those encountered during training. These changes raise a central question: \textbf{{Can a spatio-temporal forecasting model remain effective as where it is deployed, when it is deployed, and what observations are available change?}} 

Existing methods face several limitations in such vayring deployment environments. Specifically, node-specific embeddings and training-graph-dependent parameters restrict direct transfer to unseen nodes or networks. Besides, many models draw information only from a predefined set of nodes. 
Even when they dynamically adjust the importance of these nodes, they cannot seek useful information from nodes outside that set. Models may also rely heavily on observations that are consistently available during training, making accurate forecasting difficult when these observations are missing at deployment. These limitations reflect a common challenge: \textbf{{the forecasting mechanism is tightly coupled to the training environment.}}

This coupling is problematic because the information useful for forecasting is itself environment-dependent. Temporal shifts require reassessing the target’s current behavior; spatial changes may alter which nodes provide useful support; and changes in observation availability may make previously useful information unavailable. 
We refer to the information selected to support a forecast as predictive evidence, and its organization as predictive context. 
We therefore argue that \textbf{{a generalizable model should dynamically construct its predictive context from currently available observations, rather than inherit it from the training environment.}} 
This requires characterizing the target node’s current behavior, identifying where relevant evidence can be found, and organizing that evidence for the current prediction.

Long-term observations provide the basis for dynamically constructing predictive context. A short recent window captures immediate dynamics but offers limited evidence for distinguishing persistent behavior from transient fluctuations. Longer histories reveal recurring patterns and operating regimes, while providing alternative cues when recent observations are incomplete. They therefore help characterize the target node’s current behavior and identify useful predictive support in the current deployment environment. Their value thus extends beyond a longer input horizon to enabling predictive context to be reconstructed as deployment conditions change.

Building on this perspective, we propose \textbf{VaryST}, a compact framework trained from scratch for generalizable \textbf{\underline{ST}} forecasting in \textbf{\underline{Vary}}ing deployment environments. VaryST dynamically constructs target-specific predictive contexts from available long-term observations. Specifically, Long-Term Observation-Derived Representations characterize node behavior without fixed identities, while Dynamic Functional Support Node Retrieval identifies relevant support nodes beyond predefined neighborhoods. Target-Centric Hierarchical Contextual Decoding organizes recent and long-term target states, retrieved supports, and global context across multiple temporal scales and spatial scopes, providing a unified interface for inductive inference. To improve robustness to varying observation availability, Importance-Guided Observation Perturbation (IGOP) masks highly relied-upon observations during training, encouraging the model to reorganize predictive evidence. Together, these designs enable VaryST to reconstruct predictive context as deployment conditions change. Our main contributions are summarized as follows:


\begin{itemize}[leftmargin=*]
\item We identify training-environment coupling as a common challenge across temporal shifts, spatial changes, and varying observation availability, and propose \textbf{VaryST} to address it through dynamic predictive context construction.

\item We propose Long-Term Observation-Derived Representations and Dynamic Functional Support Node Retrieval to characterize nodes without fixed identities and identify relevant support beyond predefined neighborhoods, enabling adaptation to temporal shifts and spatial changes. 

\item We develop Target-Centric Hierarchical Contextual Decoding to organize multi-scale target-specific evidence through a unified inductive interface, together with Importance-Guided Observation Perturbation for robustness to varying observation availability.

\item Extensive experiments on six real-world datasets against seventeen baselines demonstrate strong in-domain forecasting, zero-/few-shot generalization across temporal and spatial changes, and robustness to unseen observation conditions. 

\end{itemize}


\section{Related Work}
\label{related_work}

\subsection{Spatio-Temporal Forecasting}
A central challenge in spatio-temporal forecasting is modeling temporal dependencies within each node and spatial dependencies across nodes. Recurrent models such as LSTM~\citep{lstm} capture temporal dynamics. DCRNN~\citep{DCRNN}, STGCN~\citep{STGCN}, Graph WaveNet~\citep{GRAPHWAVENET}, and AGCRN~\citep{AGCRN} combine recurrent or temporal convolutional modeling with graph-based spatial modeling, while ASTGCN~\citep{ASTGCN}, ASTGNN~\citep{ASTGNN} and PDFormer~\citep{PDFormer} incorporate attention to capture spatio-temporal dependencies. These models achieve strong in-domain performance. However, some rely on node-specific embeddings or adjacency parameters tied to the training network, limiting their direct application to unseen nodes or networks. 
Recently, pretrained language model (PLM)-based approaches and spatio-temporal foundation models have demonstrated strong forecasting and generalization capabilities. OFA~\citep{OFA}, ST-LLM~\citep{STLLM}, STG-LLM,~\citep{STGLLM} and STD-PLM~\citep{STD-PLM} adapt pretrained language models to spatio-temporal modeling, while UniST~\citep{yuan2024unist} learns transferable knowledge from diverse spatio-temporal data. These advances highlight the value of pretrained representations and transferable knowledge. However, such capabilities alone do not ensure that a model can identify alternative predictive evidence when node behavior changes or previously available observations become missing. 
Some studies investigate out-of-distribution (OOD) spatio-temporal forecasting to improve robustness across environments, including spatial and temporal shifts in STONE~\citep{wang2024stone} and STOP~\citep{ma2025robust}. While these studies focus on particular deployment challenges, we consider a broader perspective involving changes in spatial domains, temporal distributions, and observation availability. 

\subsection{Long-Context Spatio-Temporal Modeling}
PatchTST~\citep{patchtst}, iTransformer~\citep{iTransformer}, and SparseTSF~\citep{lin2026sparsetsf} demonstrate the value of extended temporal dependencies but do not explicitly incorporate spatial graph structure. SSTBAN~\citep{SSTBAN}, STID~\citep{shao2022spatial}, BiST~\citep{ma2025bist}, and FaST~\citep{zhao2026fast} explore efficient spatio-temporal modeling, while RAST~\citep{ruan2026retrieval} retrieves historical patterns to augment prediction. STMAE~\citep{sun2024revealing} and BigST~\citep{han2024bigst} further exploit historical representations through separate pretraining stages. These studies motivate efficient use of historical information for forecasting. 
These studies motivate efficient historical modeling. VaryST further exploits long-term observations directly to guide spatial support selection and predictive evidence organization. 

\section{Preliminaries} 
\textbf{Spatio-Temporal Data.}
We consider a spatio-temporal system defined over a spatial graph $\mathcal{G}=(\mathcal{V},\mathcal{A})$, where $\mathcal{V}=\{v_1,\ldots,v_N\}$ is the set of $N$ spatial nodes and $\mathcal{A}$ represents their spatial relations. At time $t$, the observations are denoted by $\mathbf{X}_t\in\mathbb{R}^{N\times C}$, where $C$ is the number of features. Their availability is indicated by a binary mask $\mathbf{M}_t\in\{0,1\}^{N\times C}$, where $M_{t,i,c}=1$ if the corresponding observation is available.

\textbf{Varying Deployment Environments.} We characterize deployment environments along three dimensions: temporal distributions, spatial domains, and observation availability. These dimensions may differ between training and deployment or change during deployment. 
\textit{Evolving temporal distributions} refer to changes in observation distributions over time, $P_t(\mathbf X)\neq P_{t'}(\mathbf X)$, due to seasonality, long-term trends, or changing operating regimes. 
\textit{Changing spatial domains} refer to changes in the underlying graph, $\mathcal G\neq\mathcal G'$, including the addition or removal of nodes and spatial relations, or transfer to a different network. 
\textit{Varying observation availability }refers to changes in the distribution of observation masks, $P(\mathbf M)\neq P'(\mathbf M)$, reflecting different missing rates or patterns. 

\textbf{Generalizable Spatio-Temporal Forecasting.}
Given observations $\mathbf{X}_{t-T+1:t}$ of $\mathcal{G}$ over the past \(T\) time steps and their availability masks $\mathbf{M}_{t-T+1:t}$, we aims to predict the future $Q$ steps:
\begin{equation}
    \hat{\mathbf{X}}_{t+1:t+Q}
    =
    f_\theta\left(
        \mathbf{X}_{t-T+1:t},
        \mathbf{M}_{t-T+1:t};
        \mathcal{G}
    \right).
\end{equation}
Our goal is to learn a forecasting model that generalizes beyond its training environment to deployment environments with different temporal distributions, spatial graphs, or observation availability. For spatial and temporal generalization, we consider both zero-shot forecasting without parameter updates and few-shot adaptation using limited target-environment training data. For changes in observation availability, we evaluate generalization to unseen missing rates and patterns without retraining.

\section{Methodology} 
\label{Methodology}

\begin{figure}
    \centering
    \includegraphics[width=1\linewidth]{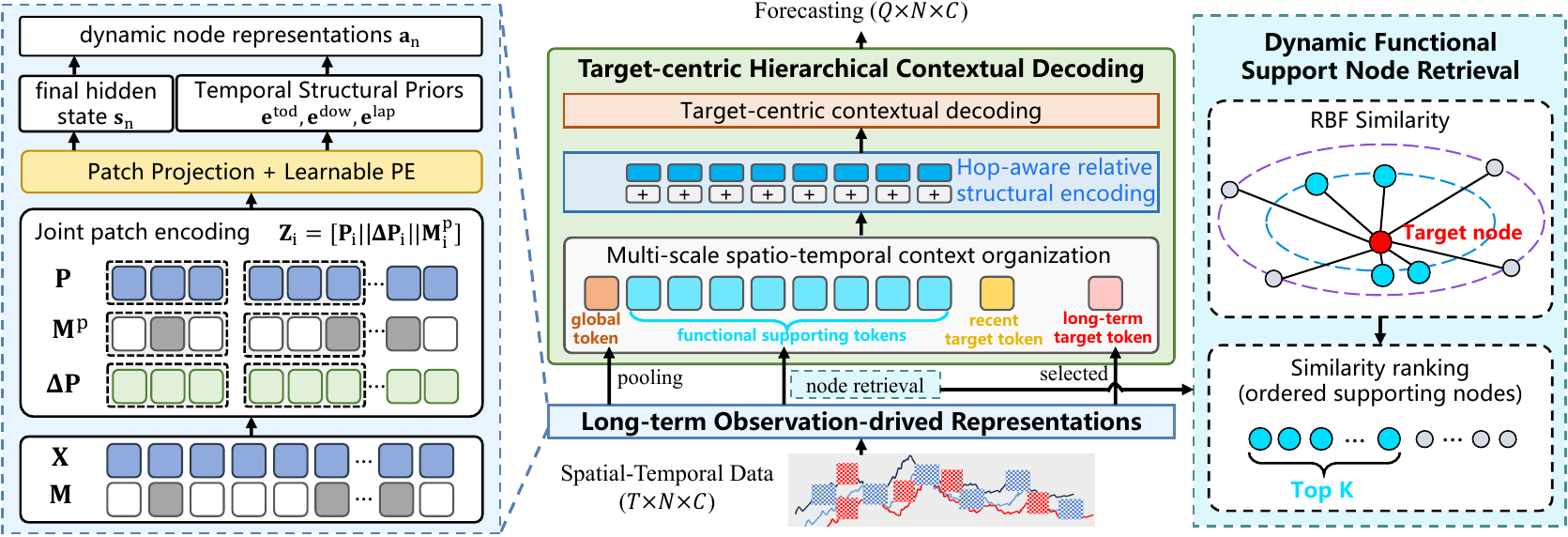}
    \caption{The architecture of VaryST. It contains Long-term Observation-derived Representations, Dynamic Functional Support Node Retrieval, and Target-centric Hierarchical Contextual Decoding.}
    \label{fig:varst_arch}
    \vspace{-10pt}
\end{figure}
We propose VaryST, a compact forecasting framework trained from scratch that dynamically selects and organizes predictive evidence for varying deployment environments. \textit{Long-Term Observation-Derived Representations} characterize each node’s current state without fixed node-specific embeddings and guide \textit{Dynamic Functional Support Node Retrieval}, allowing predictive evidence to adapt to changing spatial domains and temporal regimes. \textit{Target-Centric Hierarchical Contextual Decoding }organizes recent and long-term target states, retrieved functional supports, and global information into a compact context spanning complementary temporal scales and spatial scopes, and further incorporates target-relative structural information through hop-aware encoding. Finally, \textit{Importance-Guided Observation Perturbation} encourages the model to exploit alternative temporal and spatial evidence, improving robustness to varying observation availability.

\subsection{long-term observation-derived representations}
VaryST represents each node through its available long-term observations rather than a fixed node-specific embedding, enabling node representations to adapt to the current deployment environment.

\textbf{Linear interpolation and temporal differences.} First-order differences help characterize temporal fluctuations, but computing them directly from incomplete observations may introduce artificial changes around missing positions. Therefore, given observations $\mathbf X$ and availability mask $\mathbf M$, we first apply linear interpolation \(
\widetilde{\mathbf X}=\operatorname{Interp}(\mathbf X,\mathbf M)\), and then compute
\(\Delta\widetilde{\mathbf X}_t=\widetilde{\mathbf X}_t-\widetilde{\mathbf X}_{t-1}\), initializing the first difference to zero. 

\textbf{Joint patch encoding.} We partition \(\widetilde{\mathbf X}\), \(\Delta\widetilde{\mathbf X}\), and \(\mathbf M\) into \(T_p=\lceil T/L_p\rceil\) temporally aligned patches. For node \(v_n\), the \(i\)-th patch representation is
\(\mathbf z_{n,i}=[\mathbf P_{n,i}\|\Delta\mathbf P_{n,i}\|\mathbf M^p_{n,i}],
\) where \(\|\) denotes feature-wise concatenation.
The patch sequence is encoded via a lightweight causal Transformer with learnable temporal order embedding \(\mathbf E^{pos}\in\mathbb R^{T_p\times d}\) of patches:
\begin{equation} 
\mathbf H_n
=
\operatorname{Enc}
\left(
\operatorname{Proj}(\mathbf Z_n)
+
\mathbf E^{pos}
\right)\in\mathbb R^{T_p\times d}.
\end{equation} 
Its final hidden state \(\mathbf s_n=\mathbf H_n[T_p]\in\mathbb R^d\) summarizes all preceding historical patches and serves as the dynamic node representation.

\textbf{Temporal and structural priors.} 
We further incorporate lightweight node-ID-free temporal and structural priors, including time-of-day, day-of-week, and Laplacian positional embeddings~\citep{STD-PLM}. The resulting dynamic node representation is 
\begin{equation}
    \mathbf a_n
=\mathbf s_n
+
\mathbf e^{tod}_t
+
\mathbf e^{dow}_t
+
\mathbf e^{lap}_n.
\end{equation}
Because each $a_n$ is derived from observations rather than a learned node identity, the same encoder can construct representations for unseen nodes and update those of existing nodes as their operating regimes change. These dynamic node representations are then used for functional support retrieval and target-centric contextual decoding.

\subsection{Dynamic Functional Support Node Retrieval}


Predefined graph neighborhoods or spatial dependencies learned over fixed training nodes may become unreliable when the spatial domain changes, informative nodes lie outside the predefined neighborhood, or previously useful observations become unavailable. We therefore retrieve support nodes dynamically from their current representations rather than inheriting a fixed support set from training.

For each target node \(v_n\), VaryST measures its affinity with every candidate node \(v_j\) using their dynamic representations \(\mathbf{a}_n\) and \(\mathbf{a}_j\). Specifically, we employ an RBF similarity:
\begin{equation}
  \kappa_{n,j}=
\exp
\left(
-
\left\|
\mathbf a_n-\mathbf a_j
\right\|_2^2
\right),
\qquad j\neq n.  
\end{equation}
Because \(\mathbf{a}_n\) captures the current observation-derived state together with temporal and structural priors, \(\kappa_{n,j}\) reflects functional compatibility under the current context rather than predefined spatial proximity. We omit the RBF bandwidth because the similarity is used only for ranking, for which any fixed positive bandwidth preserves the same ordering.

The \(K\) nodes with the highest similarities form the functional supporting set of target node \(v_n\):
\begin{equation}
    \mathcal S_n
=
\operatorname{TopK}_{j\neq n}
\left(
\kappa_{n,j}
\right)
=
\{v_{n,1},\ldots,v_{n,K}\}.
\end{equation}
The retrieved nodes are ordered by decreasing similarity, and their representations are subsequently used as supporting tokens in the target-centric decoder. 

Unlike predefined graph neighborhoods, the retrieved support set adapts to current node representations and can change with temporal regimes, spatial domains, and observation availability, allowing VaryST to identify alternative predictive evidence when previously relied-upon nodes become less informative or unavailable. Since retrieval requires neither node-specific identifiers nor training-graph-specific support sets, it is naturally applicable to unseen graphs. Dynamic retrieval thus determines which nodes enter the predictive context, while the subsequent decoder determines how they influence the target.

\subsection{Target-Centric Hierarchical Contextual Decoding}
After retrieving functionally relevant support nodes, VaryST organizes predictive evidence into a target-centric hierarchical context spanning complementary temporal scales and spatial scopes.

\textbf{Multi-scale spatio-temporal context organization.}
For a target node \(v_n\), we first construct a global token by pooling the dynamic representations of all nodes:
\begin{equation}
\mathbf g
=
\operatorname{MeanPool}
(\mathbf a_1,\ldots,\mathbf a_N)
\in\mathbb R^d.
\end{equation}
To complement the long-term target representation \(\mathbf a_n\), we explicitly preserve its recent dynamics by projecting the latest \(R\) observations and first-order differences:
\begin{equation}
 \mathbf r_n
=
\operatorname{MLP}_{r}
\left(
[
\widetilde{\mathbf X}_{t-R+1:t,n}
\|
\Delta\widetilde{\mathbf X}_{t-R+1:t,n}
]
\right). 
\end{equation}
Given the retrieved supporting set \(\mathcal S_n\),
the target-specific context is organized as:
\begin{equation}
\mathbf C_n^0=
[
\underbrace{\mathbf g}_{\text{global token}}
\|
\underbrace{\mathbf a_{n,1},\ldots,\mathbf a_{n,K}}_{\text{functional supporting tokens}}
\|
\underbrace{\mathbf r_n}_{\text{recent target token}}
\|
\underbrace{\mathbf a_n}_{\text{long-term target token}}
].
\end{equation}
This hierarchy combines the target's immediate and long-term dynamics with functional spatial support and global system context, providing complementary evidence across temporal scales and spatial scopes. 

\textbf{Hop-aware relative structural encoding.} Functional retrieval identifies support nodes according to their current functional similarity, but does not distinguish their structural relations to the target. To preserve such information, we encode the target-relative graph distance of each supporting node. For \(v_{n,j}\in\mathcal{S}_n\), we compute its shortest-path hop distance to the target,
\begin{equation}
    h_{n,j}=\operatorname{HopDist}(v_n,v_{n,j}),
\end{equation}
and map the discretized distance to a learnable phase vector \(\boldsymbol{\phi}_{h_{n,j}}\in\mathbb{R}^{d/2}\). We then apply a rotary transformation to the corresponding supporting token:
\begin{equation}
    \bar{\mathbf a}_{n,j}
=
\mathbf R(\boldsymbol{\phi}_{h_{n,j}})
\mathbf a_{n,j},
\qquad j=1,\ldots,K,
\end{equation}
where \(\mathbf R(\cdot)\) applies a block-diagonal rotation to each pair of feature dimensions. Importantly, only the supporting-node tokens are structurally encoded; the global, recent-target, and long-term-target tokens remain unchanged. Since hop distance describes a target-relative relation rather than node identity or absolute graph position, the same encoding can be directly reused on unseen graphs, providing an inductive structural prior for cross-graph decoding.

\textbf{Target-centric contextual decoding.} 
The resulting context is processed by an \(L\)-layer lightweight Transformer decoder:
\begin{equation}
    \mathbf C_n^{L}=
\operatorname{Decoder}
\left(
\mathbf C_n^0
\right).
\end{equation}
We disable positional encoding in the decoder so that no sequential ordering is imposed on the heterogeneous context tokens; spatial structure is introduced exclusively through the target-relative hop encoding of support nodes. The long-term target token is placed at the final position, and its output state \(\mathbf o_n=\mathbf C_n^L[-1]\) aggregates information from the global, functional-support, and recent-target contexts. Forecasts are produced as:

\begin{equation}
    \widehat{\mathbf Y}_n
=
\operatorname{MLP}
\left(
\operatorname{LN}(\mathbf o_n)
\right),
\qquad
\widehat{\mathbf Y}_n
\in
\mathbb R^{Q\times C}.
\end{equation}

\subsection{Training with Importance-Guided Observation Perturbation}
To improve robustness to varying observation availability, we introduce Importance-Guided Observation Perturbation (IGOP) during training. Rather than masking observations randomly, IGOP deliberately removes historical patches on which the model relies most, creating harder, importance-guided perturbations that encourage the model to exploit alternative temporal and spatial evidence. 

\textbf{Patch Importance Estimation.}
Given the long-term patch representations of node \(v_n\),
\(\mathbf H_n=
[\mathbf h_{n,1},\ldots,\mathbf h_{n,T_p}],\) we use the attention scores produced by the joint patch encoder to estimate patch importance. Let
\(
\boldsymbol{\alpha}_n=
[\alpha_{n,1},\ldots,\alpha_{n,T_p}]
\), where a larger \(\alpha_{n,p}\) indicates stronger dependence on patch \(p\). These scores are used only to guide perturbation construction.

\textbf{Importance-Guided Observation Perturbation.} 
For each node, IGOP selects the \(K_m\) patches with the highest attention-derived importance scores,
\begin{equation}
    \mathcal P_n^{\mathrm{imp}}
=
\operatorname{TopK}_{p\in\{1,\ldots,T_p\}}
\left(\alpha_{n,p},K_m\right),
\end{equation}
All observations within the selected patches are masked to obtain a perturbed observation mask \(\mathbf M'\). Since each patch corresponds to a contiguous temporal interval, the perturbation naturally produces block-wise observation loss. The interpolated observations and temporal differences are then recomputed under \(\mathbf M'\) to prevent masked values from leaking into subsequent representations.

\textbf{Two-Stage Perturbation Training.}
For each mini-batch, a gradient-free probe first estimates patch importance and constructs \(\mathbf M'\). VaryST is then optimized on the perturbed history with the original forecasting target:
\begin{equation}
  \widehat{\mathbf Y}
=
f_{\theta}(\mathbf X,\mathbf M'),
\qquad
\mathcal L_{\mathrm{pred}}
=
\frac{1}{NQC}
\left\|
\widehat{\mathbf Y}-\mathbf Y
\right\|_1 .  
\end{equation}
The probe participates only in perturbation construction. No reconstruction objective is imposed on the masked patches; instead, forecasting supervision directly encourages the model to reorganize its reliance on the remaining temporal and spatial evidence.

\section{Experiments} 
\label{sec:experiments}

We evaluate the proposed VaryST on real-world traffic datasets to address the following research questions:
\textbf{RQ1: }Does VaryST remain competitive under standard in-domain forecasting across different spatio-temporal systems?
\textbf{RQ2:} Can VaryST remain effective under evolving temporal distributions?
\textbf{RQ3: }Can VaryST generalize to unseen spatial environments?
\textbf{RQ4:} Is VaryST robust to varying observation availability?
\textbf{RQ5:} How do the key designs of VaryST contribute to its effectibeness?
\textbf{RQ6:} How do model hyperparameters affect model performance?
\textbf{RQ7:} What is the computational efficiency of VaryST?

\subsection{Experimental Setup}
\begin{table}
\vspace{-20pt}
  \caption{Dataset statistics.}
  \label{tab:datasets}
  \centering
  \scriptsize
  \setlength{\tabcolsep}{3pt}
  \vspace{-10pt}
  \begin{tabular}{ccccc}
    \toprule
    Dataset & Nodes & Time Span & Interval & Usage  \\
    \midrule
    PEMS03        & 358 & 09/01/2018 \textendash{} 11/30/2018 & 5 min  & RQ1/3/5/6 \\
    PEMS04        & 307 & 01/01/2018 \textendash{} 02/28/2018 & 5 min  & RQ1/3/4 \\
    PEMS07        & 883 & 05/01/2017 \textendash{} 08/31/2017 & 5 min  & RQ1/3/7 \\
    PEMS08        & 170 & 07/01/2016 \textendash{} 08/31/2016 & 5 min  & RQ1/3/4/5/6 \\
    PM2.5         & 12  & 03/01/2013 \textendash{} 02/28/2017 & 60 min & RQ1 \\
    PEMSD3-Stream & 655 & 2011 \textendash{} 2017 & 5 min  & RQ2 \\
    \bottomrule
  \end{tabular}
  \vspace{-10pt}
\end{table}
\textbf{Dataset.} 
Table~\ref{tab:datasets} summarizes the benchmarks. PEMS03/04/07/08 cover in-domain and cross-network forecasting, with PEMS04/08 additionally evaluated under random (RM) and block (CM) missingness. PM2.5 evaluates non-traffic generalization, while the 7-year PEMSD3-Stream with an expanding network assesses long-term temporal shift. All benchmarks use a 6:2:2 chronological split to forecast 12 future steps from 2016 historical steps (168 for PM2.5), measured by MAE, RMSE, and MAPE.



\noindent \textbf{Baselines.} 
We compare VaryST with three groups of models for standard forecasting: classical spatio-temporal and time-series models, including ASTGCN~\citep{ASTGCN}, AGCRN~\citep{AGCRN}, SSTBAN~\citep{SSTBAN}, PDFormer~\citep{PDFormer}, iTransformer~\citep{iTransformer}, and RAST~\citep{ruan2026retrieval}; pre-trained LLM-based models, including OFA~\citep{OFA}, STLLM~\citep{STLLM}, STGLLM~\citep{STGLLM}, and STD-PLM~\citep{STD-PLM}; and extended-history models, where STID~\citep{shao2022spatial}, iTransformer~\citep{iTransformer}, PatchSTG~\citep{fang2025efficient}, BiST~\citep{ma2025bist}, FaST~\citep{zhao2026fast}, and SparseTSF~\citep{lin2026sparsetsf} are adapted to the same 2016-step history as VaryST. For temporal generalization, we compare with STOP~\citep{ma2025robust}, which is specifically designed for temporal distribution shifts, and STD-PLM, which reports strong generalization forecasting ability. For spatial generalization, we compare with STD-PLM, the strongest baseline that supports cross-network transfer. For missing-observation forecasting, we select FaST, STID, and BiST for their strong standard-forecasting performance and efficiency, and use linear interpolation to pre-fill their missing inputs. 


\textbf{Settings.}
All experiments are conducted on a single NVIDIA RTX A4000 GPU under PyTorch 2.11. 
VaryST is trained from scratch with the MAE loss, Muon and AdamW optimizers with learning rate 0.001, batch size 128. Unless otherwise specified, we set the history length \(T=2016\), patch length \(L_p=36\), representation dimension \(d=96\), number of retrieved support nodes \(K=64\), recent window length \(R=12\), and decoder depth \(L=3\). For PM2.5, \(T=168\) and \(L_p=12\), while the other settings remain unchanged.
Detailed settings are provided in Table~\ref{tab:default_hyperparam}. 

\subsection{Standard Forecasting Performance (RQ1)}
\begin{table*}[t]
\centering
\caption{Overall prediction performance comparison on four PEMS datasets. Models are categorized into short-term, LLM based and long-term groups. \textbf{Bold}: best, \underline{underline}: second best.}
\vspace{-10pt}
\label{tab:forecast_performance}
\setlength{\tabcolsep}{3pt}
\resizebox{\textwidth}{!}{%
\footnotesize
\begin{tabular}{l ccc ccc ccc ccc}
\toprule
\multirow{2}{*}{\textbf{Model}} 
& \multicolumn{3}{c}{\textbf{PEMS03}} 
& \multicolumn{3}{c}{\textbf{PEMS04}} 
& \multicolumn{3}{c}{\textbf{PEMS07}} 
& \multicolumn{3}{c}{\textbf{PEMS08}} \\
\cmidrule(lr){2-4} \cmidrule(lr){5-7} \cmidrule(lr){8-10} \cmidrule(lr){11-13}
& MAE & RMSE & MAPE 
& MAE & RMSE & MAPE 
& MAE & RMSE & MAPE 
& MAE & RMSE & MAPE \\
\midrule

\multicolumn{13}{l}{\textbf{Models with Short-term History} (Input Horizon = 12)} \\
\midrule
ASTGCN & 17.85 & 29.88 & 17.65\% & 22.42 & 34.75 & 15.87\% & 25.98 & 39.65 & 11.84\% & 18.86 & 28.55 & 12.50\% \\
AGCRN & 15.98 & 28.25 & 15.23\% & 19.83 & 32.26 & 12.97\% & 22.37 & 36.55 & 9.12\% & 15.95 & 25.22 & 10.09\% \\
SSTBAN & 15.90 & 26.11 & 17.27\% & 18.88 & 31.10 & 12.56\% & 20.17 & 33.45 & 8.93\% & 14.39 & 24.19 & 10.06\% \\
PDFormer & 14.74 & 25.59 & 15.35\% & 18.31 & 29.97 & 12.10\% & 19.83 & 32.87 & 8.53\% & 13.58 & 23.51 & 9.05\% \\
iTransformer & 19.48 & 31.20 & 17.84\% & 22.56 & 35.21 & 16.29\% & 24.70 & 37.91 & 11.40\% & 20.05 & 31.90 & 11.99\% \\
RAST & 15.20 & 25.81 & 16.12\% & 18.39 & 29.93 & 12.43\% & 19.52 & 32.73 & 8.23\% & 14.16 & 23.33 & 9.27\% \\

\addlinespace[0.8em]
\multicolumn{13}{l}{\textbf{Pre-trained LLM-based Models}} \\
\midrule
OFA & 20.96 & 33.43 & 19.11\% & 27.37 & 42.99 & 17.97\% & 30.53 & 47.51 & 12.98\% & 21.89 & 34.63 & 13.30\% \\
STLLM & 17.25 & 27.25 & 22.96\% & 19.00 & 30.35 & 13.55\% & 21.48 & 34.07 & 10.20\% & 14.67 & 23.50 & 10.63\% \\
STGLLM & 15.26 & \underline{24.11} & 15.73\% & 20.00 & 32.11 & 13.69\% & 21.98 & 35.02 & 9.72\% & 15.53 & 24.74 & 10.15\% \\
STD-PLM & \underline{14.59} & 25.36 & 14.92\% & \underline{18.16} & 30.21 & \underline{11.89\%} & 19.25 & 32.84 & 8.06\% & \underline{13.31} & 23.19 & \underline{8.84}\% \\
\addlinespace[0.8em]
\multicolumn{13}{l}{\textbf{Models with Long-term History}} \\
\midrule
STID$^\dagger$ & 15.56 & 25.37 & 15.25\% & 18.40 & 30.10 & 13.19\% & 19.11 & \underline{31.71} & 8.45\% & 13.42 & \underline{22.48} & 9.37\% \\
iTransformer$^\dagger$ & 17.25 & 27.70 & 18.62\% & 20.82 & 32.79 & 14.42\% & 22.19 & 34.77 & 10.00\% & 15.31 & 23.94 & 10.61\% \\
PatchSTG$^\dagger$ & 16.92 & 28.30 & 17.13\% & 18.67 & 30.44 & 13.10\% & 20.00 & 32.29 & 9.51\% & 13.73 & 23.03 & 9.58\% \\
BiST$^\dagger$ & 15.01 & 24.64 & \underline{14.87\%} & 18.46 & \underline{29.86} & 13.22\% & 19.55 & 32.53 & 8.87\% & 13.71 & 22.79 & 9.57\% \\
FaST$^\dagger$ & 16.68 & 26.98 & 16.12\% & 18.57 & 30.10 & 12.85\% & \underline{18.92} & 31.92 & \underline{8.05}\% & 14.22 & 23.21 & 10.31\% \\
SparseTSF$^\dagger$ & 19.87 & 33.24 & 22.40\% & 19.91 & 31.94 & 14.51\% & 21.43 & 34.66 & 9.56\% & 16.12 & 25.74 & 11.52\% \\

\textbf{VaryST} & \textbf{13.95} & \textbf{22.73} & \textbf{13.59\%} & \textbf{17.56} & \textbf{29.35} & \textbf{11.81\%} & \textbf{18.21} & \textbf{31.05} & \textbf{7.64\%} & \textbf{12.89} & \textbf{21.99} & \textbf{8.50\%} \\
\bottomrule
\end{tabular}
}
\vspace{-10pt}
\begin{minipage}{\textwidth}
\footnotesize
\raggedright
$^\dagger$ Models adapted to long-term inputs by us for fair comparison (2016 steps). 
\end{minipage}
\vspace{-2mm}
\end{table*}

\begin{wraptable}[14]{r}{0.48\textwidth}
\vspace{-3mm}
  \setlength{\abovecaptionskip}{0pt}
  \caption{Performance comparison on PM2.5.}
  \label{tab:forecast_performance_pm25}
  \centering
  \scriptsize
  \setlength{\tabcolsep}{4pt}
  \begin{tabular}{l ccc}
    \toprule
    \textbf{Model} & MAE & RMSE & MAPE \\
    \midrule
    ASTGCN                  & 27.42 & 48.70 & 52.4\% \\
    AGCRN                   & 28.20 & 49.57 & \underline{50.1\%} \\
    SSTBAN                  & 30.22 & 53.13 & 53.1\% \\
    PDFormer                & \underline{26.65} & \textbf{47.08} & 51.5\% \\
    iTransformer            & 29.56 & 51.80 & 55.9\% \\
    STD-PLM                 & 27.55 & 48.88 & 53.3\% \\
    STID$^\dagger$          & 31.22 & 54.88 & 64.1\% \\
    iTransformer$^\dagger$  & 29.97 & 50.52 & 58.5\% \\
    BiST$^\dagger$          & 28.11 & 47.28 & 54.5\% \\
    FaST$^\dagger$          & 30.66 & 51.99 & 59.2\% \\
    SparseTSF$^\dagger$     & 39.64 & 62.34 & 85.2\% \\
    \midrule
    \textbf{VaryST}         & \textbf{26.44} & \underline{47.18} & \textbf{49.91\%} \\
    \bottomrule
  \end{tabular}
  \vspace{1pt}
  
  \centering\scriptsize $^\dagger$Models adapted to long-term inputs (168 steps).
  \vspace{-5pt}
\end{wraptable}

Table~\ref{tab:forecast_performance} reports 12-step forecasting results on four PEMS benchmarks. Overall, long-history models outperform short-history baselines (e.g., PDFormer, RAST) and LLM-based methods (STD-PLM). VaryST achieves the best performance across all four datasets, reducing MAE by 3.16\%--4.39\%, RMSE by 1.71\%--5.72\%, and MAPE by 0.67\%--8.61\% over the strongest baselines. Its advantage over competitive long-history models (STID, FaST,BiST) demonstrates that simply feeding longer sequences is insufficient; VaryST crucially leverages long-term observations to derive dynamic node states and retrieve functional supports.

Table~\ref{tab:forecast_performance_pm25} evaluates generalization beyond traffic on PM2.5, which features hourly observations across only 12 stations without an explicit graph. VaryST achieves the best MAE and MAPE alongside a competitive RMSE, demonstrating that dynamic target-centric context construction generalizes effectively to non-traffic domains.

\subsection{Temporal Generalization (RQ2)}
\label{subsec:temporal}

\begin{figure}[t]
    \centering
    \begin{subfigure}[t]{0.30\linewidth}
        \centering
        \includegraphics[width=\linewidth]{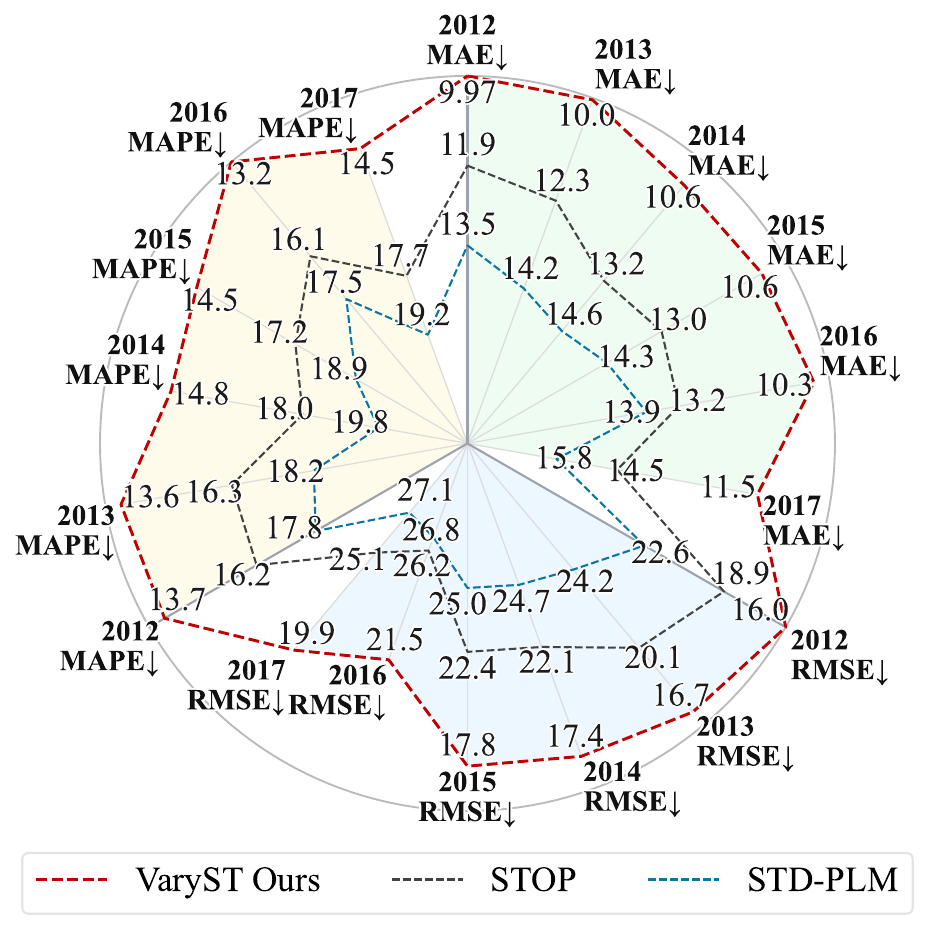}
        \caption{Per-year zero-shot metrics on PEMSD3-Stream (2012--2017).}
        \label{fig:ood_time}
    \end{subfigure}
    \hfill
    \begin{subfigure}[t]{0.34\linewidth}
        \centering
        \includegraphics[width=\linewidth]{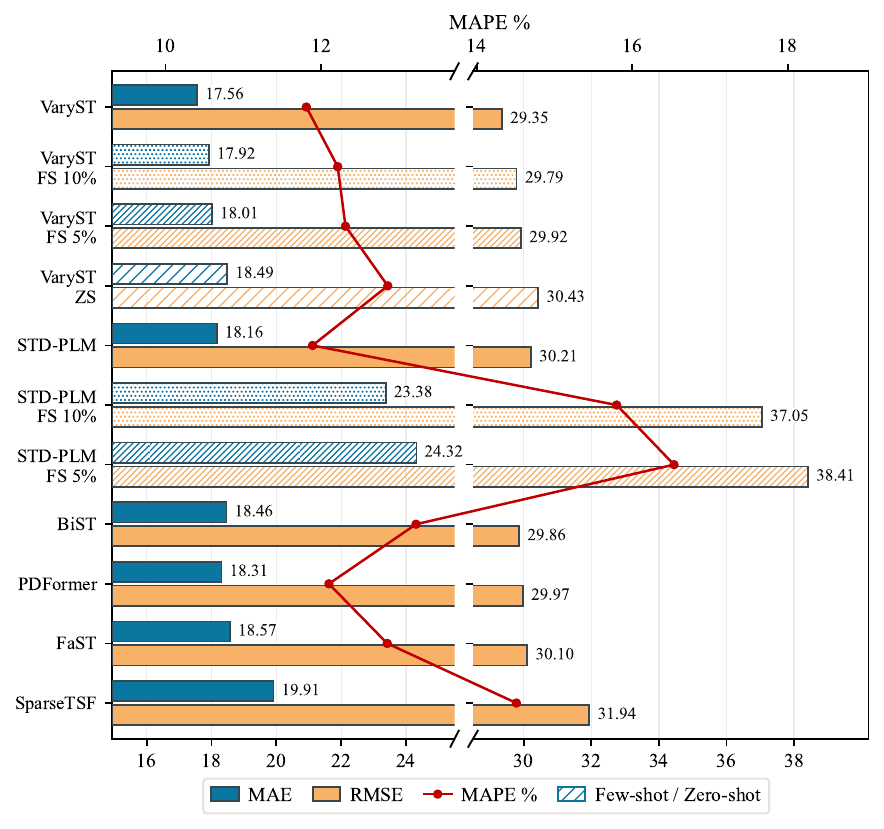}
        \caption{Zero-shot vs.\ in-domain MAE on PEMS04.}
        \label{fig:ood_spatial}
    \end{subfigure}
    \hfill
    \begin{subfigure}[t]{0.32\linewidth}
        \centering
        \includegraphics[width=\linewidth]{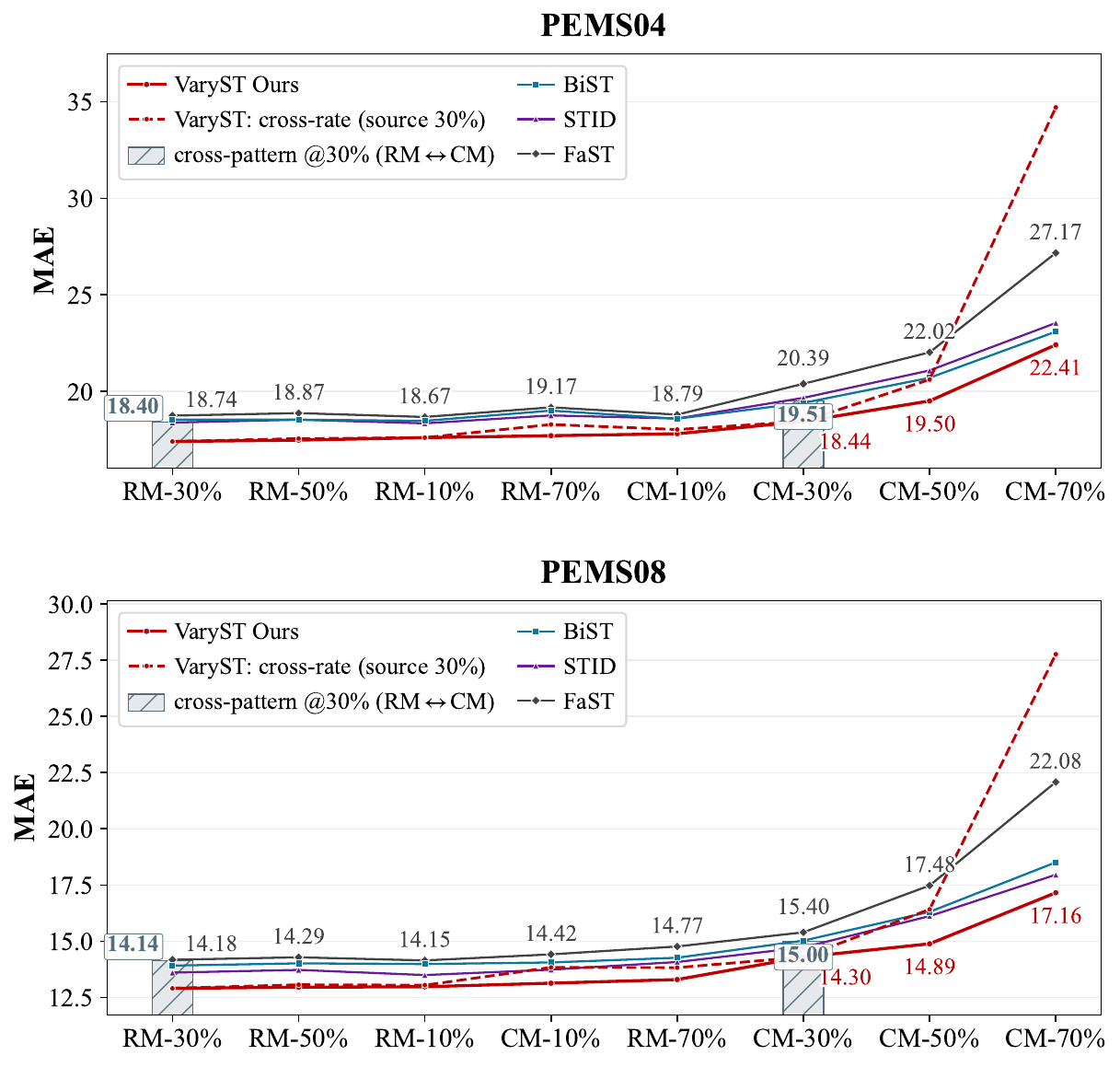}
        \caption{MAE under RM and CM at 10--70\% rates.}
        \label{fig:ood_missing}
    \end{subfigure}
    \caption{Distribution-shift robustness of VaryST.}
    \vspace{-15pt}
    \label{fig:ood_visualization}
\end{figure}

To evaluate temporal generalization, models are trained only on 2011 data of PEMSD3-Stream and tested across 2012--2017 without updates (zero-shot transfer). As shown in Figure~\ref{fig:ood_time}, VaryST consistently outperforms STOP—the strongest temporal-robust baseline—across all years, reducing MAE by 16.22\%--21.97\%, RMSE by 15.34\%--21.27\%, and MAPE by 15.43\%--18.08\%. While STOP seeks invariant representations via distributionally robust optimization, VaryST adapts directly to evolving dynamics by deriving node states and dynamically retrieving functional supports from current long-term observations.



\subsection{Spatial Generalization (RQ3)}
\label{subsec:spatial}
We evaluate spatial generalization via cross-network transfer across four PEMS benchmarks under zero-shot and few-shot (1\%, 5\%, 10\% target data) settings (full 12 directions in Appendix~\ref{app:spatial_ood}).
Figure~\ref{fig:ood_spatial} highlights results on PEMS04. Under zero-shot transfer, VaryST degrades by only 3.68\%/5.30\%/8.89\% (RMSE/MAE/MAPE) relative to its in-domain reference, whereas STD-PLM degrades by
52.36\%/63.86\%/91.33\%. This gap stems from VaryST constructing representations and functional supports dynamically from current observations rather than static graph topology. With few-shot adaptation, VaryST with 5\%--10\% target data already surpasses competitive full-data in-domain models
and consistently outperforms STD-PLM across all budgets.

\subsection{Robustness to Observations Availability(RQ4)}

We evaluate cross-network transfer across four PEMS benchmarks under zero-shot and few-shot (1\%, 5\%, 10\%) settings (Appendix~\ref{app:spatial_ood}). On PEMS04 (Figure~\ref{fig:ood_spatial}), zero-shot VaryST degrades by only
3.68\%/5.30\%/8.89\% (RMSE/MAE/MAPE) relative to in-domain, whereas STD-PLM degrades by 52.36\%/63.86\%/91.33\%, as VaryST dynamically constructs spatial supports from observations rather than static topology. With 5\%--10\% target
data, VaryST already surpasses full-data in-domain baselines and outperforms STD-PLM across all budgets.

\subsection{Ablation Studies (RQ5)}
\begin{figure}[t]
    \centering
    \begin{subfigure}[t]{0.48\linewidth}
        \centering
        \includegraphics[width=\linewidth]{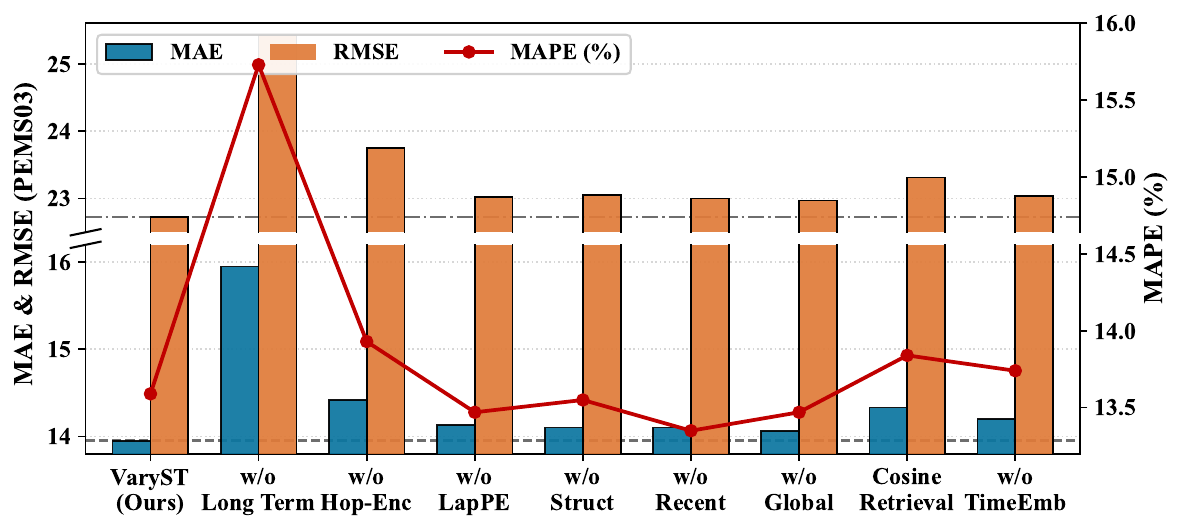}
        \caption{Ablation study on PEMS03.}
        \label{fig:ablation_pems03}
    \end{subfigure}
    \hfill
    \begin{subfigure}[t]{0.4\linewidth}
        \centering
        \includegraphics[width=\linewidth]{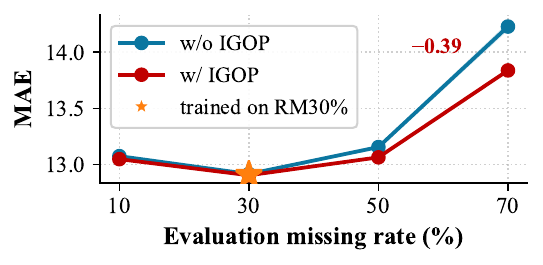}
        \caption{IGOP ablation under missing-rate shift. } 
        \label{fig:igop_transfer_ablation}
    \end{subfigure}
    
    
    \begin{subfigure}[t]{0.48\linewidth}
        \centering
        \includegraphics[width=\linewidth]{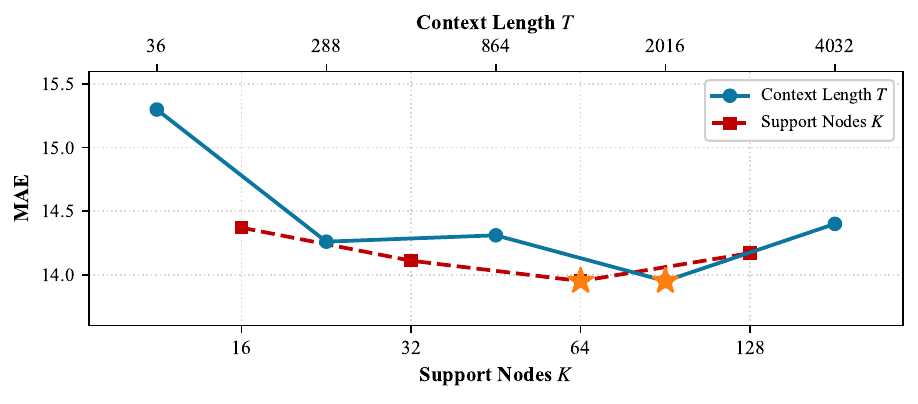}
        \caption{Hyperparameter sensitivity on PEMS03.} 
        \label{fig:hyperparam_pems03}
    \end{subfigure}
    \hfill 
    \begin{subfigure}[t]{0.4\linewidth}
        \centering
        \includegraphics[width=\linewidth]{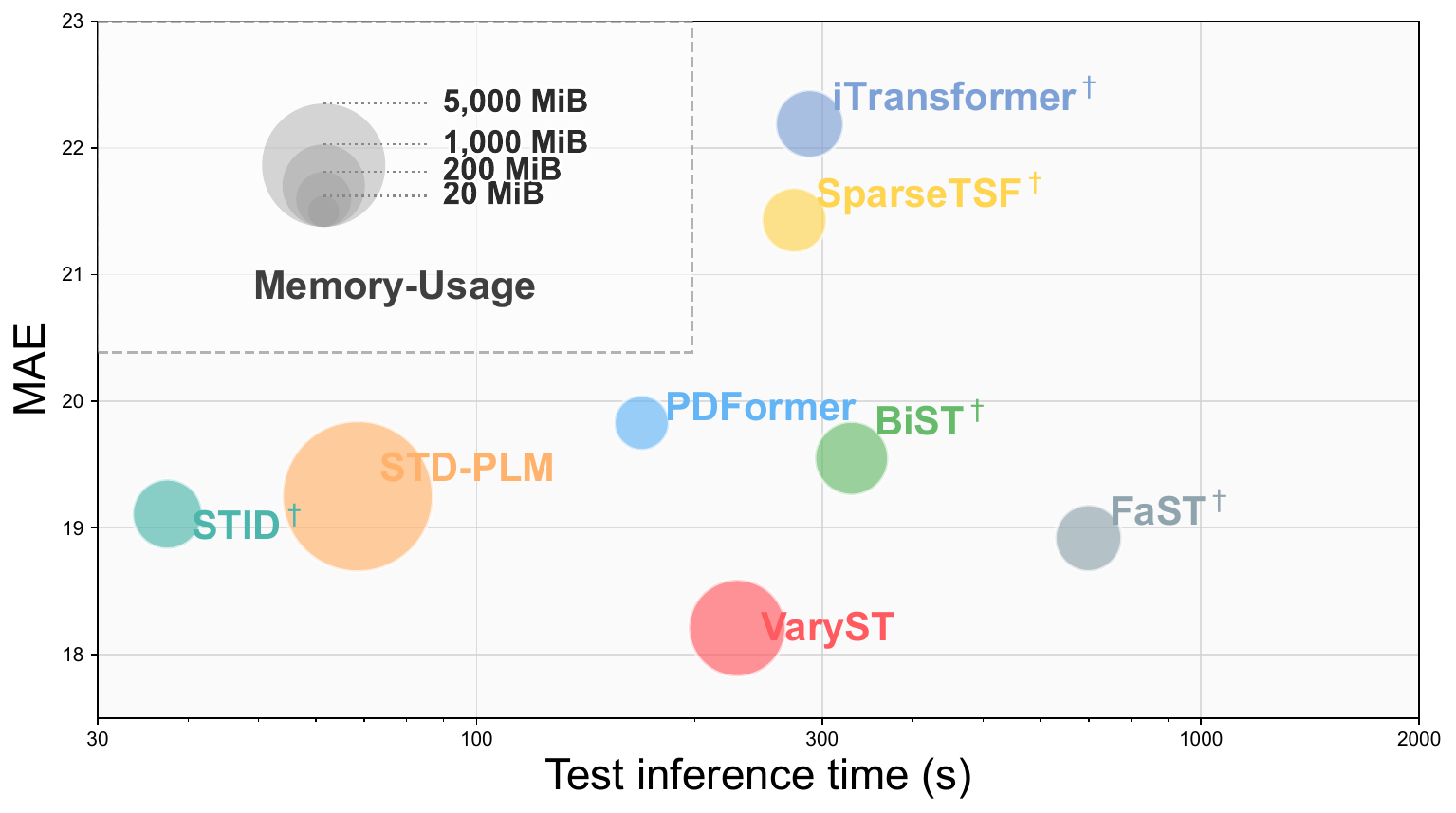}
        \caption{Efficiency on PEMS07. } 
        \label{fig:inference_efficiency}
    \end{subfigure}
    
    \caption{Ablation, hyperparameter and efficiency studies.}
    \label{fig:ablation_03_08miss}
    \vspace{-4mm}
\end{figure}

\textbf{Ablation Study} On PEMS03 (Figure~\ref{fig:ablation_pems03}), removing long-term history increases MAE from
13.95 to 15.95, validating the role of long observations in dynamic node representations. For spatial context
construction, replacing RBF similarity with cosine raises MAE to 14.33, while removing hop-aware encoding raises it to
14.42, justifying our affinity metric and target-relative structural design. For perturbation training, IGOP trained
at a 30\% missing rate reduces MAE by 0.39 at a 70\% test missing rate on PEMS08 without retraining
(Figure~\ref{fig:igop_transfer_ablation}). Full results are in Appendix~\ref{app:full_ablation}.



\textbf{Hyperparameter Sensitivity} We evaluate context length $T\in\{36,288,864,2016,4032\}$ and support count
$K\in\{16,32,64,128\}$ (Figure~\ref{fig:hyperparam_pems03}). Extending history from 36 to the default 2016 steps
reduces MAE by 8.8\% on PEMS03 and 12.8\% on PEMS08. For support retrieval, increasing $K$ from 16 to the default 64
lowers MAE by 2.9\% on PEMS03, confirming $K=64$ as an optimal trade-off. Full results are in
Appendix~\ref{app:hyperparam_full}.

\subsection{Efficiency Analysis (RQ7)}
\label{sec:efficiency_analysis}


Figure~\ref{fig:inference_efficiency} evaluates inference efficiency on PEMS07 (883 nodes). VaryST achieves the lowest MAE (18.21), reducing error by 3.8\% with faster inference than FaST$^\dagger$, while outperforming STD-PLM with substantially lower memory footprint. This efficiency stems from our compact predictive context: long-context encoding compresses each node's history into a single representation, limiting the decoder to process only $K+3$ tokens per target.

\vspace{-2mm}
\section{Conclusion}
\vspace{-2mm}

This paper presents VaryST, which addresses generalizable spatio-temporal forecasting by dynamically constructing
predictive contexts. Trained from scratch, VaryST demonstrates superior in-domain accuracy, robust zero- and few-shot
generalization across spatial and temporal shifts, and resilience to varying observation availability.

\section*{Reproducibility statement}
All datasets are public benchmarks (Table~\ref{tab:datasets}). The training scripts, together with the complete command-line configuration of every experiment, will be released upon publication.



\bibliography{iclr2027_conference}
\bibliographystyle{iclr2027_conference}

\newpage
\appendix
\raggedbottom
\section*{Appendix}
\textbf{Roadmap.} This appendix provides supplementary methodology, comprehensive experimental setups, and complete empirical results. Specifically, Appendix~\ref{app:pseudocode} presents the pseudocode for the training and inference pipelines of VaryST, along with a complexity analysis of each component. Appendix~\ref{app:setup} elaborates on dataset statistics, hyperparameter configurations, evaluation metrics, and implementation details. Due to space limitations in the main paper, Appendix~\ref{app:results} details the full experimental results, covering temporal and spatial generalization, missing-observation forecasting under rate and pattern shifts, ablation studies, and hyperparameter sensitivity analyses.

\section{Algorithms}
\subsection{Pseudo Code}
\label{app:pseudocode}
\label{app:algo}

\begin{algorithm}[H]
\caption{VaryST training and inference. Lines 3--6 run only during training; at
inference the forward pass executes once with $\mathbf M'=\mathbf M$.}
\label{alg:varyst}
\begin{algorithmic}[1]
\Require history $\mathbf X_{t-T+1:t}$, mask $\mathbf M$, graph $\mathcal G$, horizon $Q$;
patch length $L_p$, number of supports $K$, recent steps $R$, masked patches $K_m$
\Ensure prediction $\widehat{\mathbf Y}_{t+1:t+Q}$
\Statex \textit{// Stage 1: importance-guided probe (training only)}
\State $\mathbf M'\gets\mathbf M$
\If{training}
  \State $\boldsymbol\alpha_n\gets$ patch-attention scores of the long-term encoder \Comment{gradient-free probe}
  \State $\mathcal P^{\mathrm{imp}}_n\gets\operatorname{TopK}(\boldsymbol\alpha_n,K_m)$;\quad
         $\mathbf M'\gets\operatorname{Mask}(\mathbf M,\mathcal P^{\mathrm{imp}})$
\EndIf
\Statex \textit{// Stage 2: long-term dynamic representations}
\State $\widetilde{\mathbf X}\gets\operatorname{Interp}(\mathbf X,\mathbf M')$;\quad
       $\Delta\widetilde{\mathbf X}\gets$ first-order differences of $\widetilde{\mathbf X}$
\For{$n=1,\dots,N$}
  \State $\mathbf Z_n\gets[\operatorname{Patch}(\widetilde{\mathbf X}_n)\,\|\,\operatorname{Patch}(\Delta\widetilde{\mathbf X}_n)\,\|\,\operatorname{Patch}(\mathbf M'_n)]$
  \State $\mathbf a_n\gets\operatorname{Enc}(\operatorname{Proj}(\mathbf Z_n)+\mathbf E^{pos})[T_p]
         +\mathbf e^{tod}_t+\mathbf e^{dow}_t+\mathbf e^{lap}_n$
\EndFor
\Statex \textit{// Stage 3: functional support retrieval and target-centric decoding}
\State $\mathbf g\gets\operatorname{MeanPool}(\mathbf a_1,\dots,\mathbf a_N)$
\For{each target node $v_n$}
  \State $\mathcal S_n\gets\operatorname{TopK}_{j\neq n}\bigl(\exp(-\|\mathbf a_n-\mathbf a_j\|_2^2)\bigr)$;\quad
         $\mathbf r_n\gets\operatorname{MLP}_r\bigl([\widetilde{\mathbf X}_{t-R+1:t,n}\,\|\,\Delta\widetilde{\mathbf X}_{t-R+1:t,n}]\bigr)$
  \State $\mathbf C_n^0\gets[\mathbf g\,\|\,\mathbf a_{n,1},\dots,\mathbf a_{n,K}\,\|\,\mathbf r_n\,\|\,\mathbf a_n]$
  \State $\mathbf C_n^L\gets\operatorname{Decoder}(\mathbf C_n^0)$ with rotated keys
         $\widetilde{\mathbf k}_{n,j}=\mathbf R(\boldsymbol\varphi_{h_{n,j}})\mathbf k_j$, $v_j\in\mathcal S_n$
  \State $\mathbf o_n\gets\mathbf C_n^L[-1]$;\quad
         $\widehat{\mathbf Y}_n\gets\operatorname{MLP}(\operatorname{LN}(\mathbf o_n))$
\EndFor
\Statex \textit{// Stage 4: forecasting loss}
\If{training}
  \State $\theta\gets\theta-\eta\nabla_\theta\,\mathcal L_{\mathrm{pred}}$,\quad
         $\mathcal L_{\mathrm{pred}}=\tfrac{1}{NQC}\bigl\|\widehat{\mathbf Y}-\mathbf Y\bigr\|_1$
\EndIf
\end{algorithmic}
\end{algorithm}

\subsection{Complexity Analysis}
\label{app:complexity}

\paragraph{Notation.}
Let $N$ be the number of nodes, $B$ the number of sliding windows per training iteration, and $P = BN$ the number of target instances (one per window--node pair). Let $T$ and $C$ denote the input length and the number of input features, $u$ the unit-window length and $U = T/u$ the number of unit windows, $d$ the token width of the temporal encoder, $D$ the backbone width, $H$ the number of backbone layers, and $K$ the number of retrieved support nodes. With the global token and the recent prompt enabled, each target instance is decoded from a sequence of length $\ell = K + 3$.
\paragraph{Temporal encoding.}
Each window-to-node pair is encoded independently: the unit projection (two-layer MLP over each unit window) costs $O(TCd)$ per pair, the self-attention over the $U$ unit tokens costs $O(U^{2}d + Ud^{2})$, the single-query readout attention costs $O(Ud^{2})$, and the output projection costs $O(dD)$. With $U$ fixed ($U{=}56$ in our experiments), the total per iteration is $O\!\big(BN\,(TCd + Ud^{2})\big)$, linear in $B$, $N$, and $T$.
\paragraph{Support Node retrieval.}
For every target instance, similarity to all $N$ candidate nodes of its window is one $D$-dimensional inner product, followed by a top-$K$ selection, giving $O(BN^{2}D)$ per iteration. This is the only step super-linear in $N$. It is parameter-free and lies outside the gradient graph: candidate keys are detached, and the top-$K$ indices serve only as gather indices, so the step incurs no backward cost. Gradients reach the retrieved nodes through their \emph{values}, gathered from the gradient-carrying node states.
\paragraph{Context assembly and decoding.} Gathering the $K$ support tokens, the additive structural and temporal encodings, and the recent prompt token cost $O(BNKD)$ in total. The backbone then processes each target instance as a sequence of length $\ell = K + O(1)$, \emph{independent of $N$}: a Transformer layer costs $O(\ell^{2}D + \ell D^{2})$ per instance, and the hop-bucket key rotation multiplies the attention-score computation only by the constant number of buckets ($10$). Over all $P = BN$ instances and $H$ layers, decoding costs $O\!\big(BN \cdot H\,(\ell^{2}D + \ell D^{2})\big)$. Graph-dependent quantities (Laplacian eigenvectors, hop buckets, and the hop matrix) are computed once per graph, outside the training loop.
\paragraph{Where the quadratic term lives.} A graph Transformer attending over all $N$ nodes pays $O(N^{2}D)$ \emph{inside every backbone layer}, in both the forward and the backward pass, and materialises $N \times N$ attention scores in memory. In Methodology, the quadratic term appears exactly once, in the retrieval matmul above: a single parameter-free, gradient-free matrix product per iteration. All gradient-carrying computation---temporal encoding and the backbone---scales linearly with $N$ at fixed $\ell$, and the attention footprint per instance is $\ell^{2}$ rather than $N^{2}$.
\section{Experiment Details}
\label{app:setup}
\subsection{Dataset Statistics}

All six datasets are public benchmarks whose statistics are summarized in Table~\ref{tab:datasets} of the main text. The traffic datasets span 170 to 883 nodes at 5-minute sampling. The air quality dataset PM2.5 has 12 stations at hourly sampling.

\paragraph{PEMSD3-Stream protocol.}
Per-year statistics are listed in Table~\ref{tab:pemsd3_yearly_stats}. We follow the PEMSD3-Stream setting of STOP~\citep{ma2025robust}: PeMS District~3 flow over 2011--2017, one slice per year, and each year an independent graph whose station count grows from 655 to 871 as new sensors join. Training and validation use 2011 alone with the window indices released by STOP, the validation split being disjoint from training and used for early stopping, so no window of a later year is seen before testing. Testing uses the held-out windows of a year: 2011 gives the in-domain reference, while 2012--2017 are evaluated zero-shot with the 2011 checkpoint, without retraining. As in STOP, raw flow of zero marks a missing station and is excluded from the loss and the metrics, so all reported numbers are computed on observed positions.

\begin{table}[!htp]
  \caption{Per-year statistics of PEMSD3-Stream.}
  \label{tab:pemsd3_yearly_stats}
  \centering
  \small
  \setlength{\tabcolsep}{3pt}
  \begin{tabular}{cccc}
    \toprule
    Year & Nodes & Time Span & Interval \\
    \midrule
    2011 & 655 & 07/10/2011 \textendash{} 08/31/2011 & 5 min \\
    2012 & 715 & 07/10/2012 \textendash{} 08/31/2012 & 5 min \\
    2013 & 786 & 07/10/2013 \textendash{} 08/31/2013 & 5 min \\
    2014 & 822 & 07/10/2014 \textendash{} 08/31/2014 & 5 min \\
    2015 & 834 & 07/10/2015 \textendash{} 08/31/2015 & 5 min \\
    2016 & 850 & 07/10/2016 \textendash{} 08/31/2016 & 5 min \\
    2017 & 871 & 07/10/2017 \textendash{} 08/31/2017 & 5 min \\
    \bottomrule
  \end{tabular}
  \vspace{-5pt}
\end{table}

\subsection{Hyperparameters}

\begin{table}[H]
\centering
\caption{Default hyperparameters for the VaryST framework.}
\label{tab:default_hyperparam}
\setlength{\tabcolsep}{5pt}
\footnotesize
\begin{tabular}{llc}
\toprule
\textbf{Hyperparameter} & \textbf{Description} & \textbf{Value or Choices} \\
\midrule
sample\_len & Input history length $T$ & 2016 (168 on PM2.5) \\
output\_len & Prediction horizon $Q$ & 12 \\
unit\_len & Patch length $L_p$ & 36 (12 on PM2.5) \\
train\_ratio / val\_ratio & Chronological train/validation split & 0.6 / 0.2 \\
\midrule
token\_dim & Temporal Patch Token dimension $d$ & 96 \\
emb\_dim & backbone dim $D$ & 96 \\
n\_head & backbone attention heads & 4 \\
backbone\_layers & Number of backbone layers & 3 \\
spatial\_pe & Structural encodings & lap, hop\_rope \\
node\_pe\_k & Laplacian PE eigenvectors & 16 \\
n\_aux & Number of support nodes retrieved per target $K$ & 64 \\
use\_time\_emb & Time-of-day / day-of-week embeddings & True \\
use\_global\_token & Global mean-pooled token & True \\
recent\_len & Recent steps in the prompt token & 12 \\
norm\_mode & Input normalization & RevIN \\
\midrule
loss\_type & Training loss & MAE \\
optimizer & Optimizer & Muon + AdamW \\
lr & Learning rate & 0.001 \\
weight\_decay & Weight decay & 0.0 \\
dropout & Dropout & 0.05 \\
batch\_size & The batch size for training & 128 \\
lr\_patience & Learning-rate decay patience & 5 \\
patience & Early-stopping patience & 11 \\
epoch & Maximum training epochs & 500 \\
\bottomrule
\end{tabular}
\end{table}

Table~\ref{tab:default_hyperparam} lists the default configuration shared by all PEMS experiments. On PM2.5, the history length and the patch length, are changed to 168 and 12 respectively, as marked in the table. The sensitivity of context length, patch length, and the number of retrieved support nodes is studied in Section~\ref{app:hyperparam_full}.

\subsection{Metrics}
Throughout this paper, we use the following metrics to evaluate performance:

\textbf{MAE (Mean Absolute Error):} Represents the average absolute difference between the predicted and actual values, providing an interpretable measure of average error magnitude.
\begin{equation}
\mathrm{MAE} = \frac{1}{n}\sum_{i=1}^{n} |y_i - \hat{y}_i|,
\end{equation}
where $y_i$ and $\hat{y}_i$ denote the ground truth and predicted values, respectively, and $n$ is the number of evaluated data points.

\textbf{RMSE (Root Mean Squared Error):} The square root of the mean squared difference between the predicted and actual values, expressed in the same units as the original data, and more sensitive to large deviations than MAE.
\begin{equation}
\mathrm{RMSE} = \sqrt{\frac{1}{n}\sum_{i=1}^{n} (y_i - \hat{y}_i)^2}.
\end{equation}

\textbf{MAPE (Mean Absolute Percentage Error):} Expresses errors as a percentage of the actual values, offering a scale-independent metric that facilitates comparisons across datasets.
\begin{equation}
\mathrm{MAPE} = \frac{100}{n}\sum_{i=1}^{n} \left|\frac{y_i - \hat{y}_i}{y_i}\right|.
\end{equation}

All three metrics are computed only at observed positions, and reported values are averaged over all 12 prediction horizons; MAPE is expressed in percent. For all metrics, lower values indicate better performance.

\subsection{Implementation Details}

\subsubsection{Hardware and Environment}
All experiments run on a single NVIDIA RTX A4000 GPU under PyTorch 2.11. With early stopping, training completes in 3.5 hours on PEMS03, 4.2 hours on PEMS04, 4.4 hours on PEMS07, 3.8 hours on PEMS08, 2.7 hours on the PEMSD3-Stream train year, and 44 minutes on PM2.5, where the wall-clock covers the full training loop including validation and test passes. Peak training memory stays within 0.9--2.9 GB on the traffic benchmarks and below 0.1 GB on PM2.5.

\subsubsection{Data Splitting and Leakage Prevention}
Every dataset is split chronologically into training, validation, and test portions in a 6:2:2 ratio, with no shuffling across time. Zero-shot evaluation applies a model trained fully on the source dataset directly to the target dataset, without updating target parameters. Few-shot transfer fine-tunes on 1\%, 5\%, or 10\% of the target training data, matching Table~\ref{tab:transfer_fewshot}. The fraction is taken over training windows, each retaining the full 2016-step input that ends exactly where its 12-step target begins, rather than over the raw time span, which for our splits is shorter than one window and would leave nothing to train on; the retained windows are evenly spaced across the training portion, so a 1\% run performs roughly 1\% of the parameter updates of a full run. The same 2016-step input length also fixes the evaluation span: because a window's target can only start where its input ends, the validation and test boundaries are shifted back by 2016 steps, so that the targets of the test windows tile the entire test portion from its first step to its last and every test step is scored. A model that under- or over-fits the windows it was fine-tuned on is therefore judged on the whole test span, never on a span as narrow as its few-shot subset. VaryST carries no node identity: representations are derived online from current observations, and its structural priors are relative relations only, namely hop distances and Laplacian eigenvectors. No entity information from a training graph therefore survives transfer. In the missing-observation experiments the input mask coincides with the loss mask, so the loss is computed only at observed positions; see the next subsubsection.

\subsubsection{Baseline Adaptation and Missing-Input Protocol}
\label{app:baseline_adaption}
\label{app:missing_protocol}
Dagger$\dagger$-marked baselines were adapted by us to the same long-term input as VaryST, namely 2016 steps on the traffic benchmarks, while the PM2.5 comparison uses the 168-step input as stated in its table footnote. These adaptations change only the input interface, with no architectural modification.
PatchSTG additionally requires sensor latitude and longitude metadata to construct its geospatial partitioning, which the PEMS benchmarks do not provide; we therefore synthesize virtual coordinates for each dataset so that PatchSTG operates with its original geo-partitioning intact. 
Under missing observations the baselines do not consume observation masks, so their inputs are pre-filled by linear interpolation. VaryST instead consumes the mask natively in its encoder, so its representations are conditioned on which observations are available.

\section{Full Experiment Results}
\label{app:results}

\subsection{Full Temporal Generalization Results}

Models are trained and validated on 2011 alone and evaluated zero-shot on each year from 2012 to 2017 as new sensors join the network; metrics are computed only at observed positions and averaged over all 12 prediction horizons. The six-year averages are summarized in Table~\ref{tab:pemsd3_pm25} and the per-year breakdown in Table~\ref{tab:pemsd3_pm25_yearly}. VaryST attains the best MAE, RMSE, and MAPE in every year, reducing the six-year average MAE from 13.01 to 10.49. VaryST improves over STOP, the strongest baseline, by 19.4\%. The gap follows from observation-derived node representations recomputed from current observations, and from retrieval that reselects supporting evidence as temporal patterns drift. The per-year curves appear in Figure~\ref{fig:ood_time} of the main text.
%
\begin{table}[!htp]
  \vspace{-10pt}
  \caption{Average Zero-shot performance on PEMSD3-Stream.}
  \label{tab:pemsd3_pm25}
  \centering
  \small
  \setlength{\tabcolsep}{6pt}
  \begin{tabular}{l ccc}
    \toprule
    \textbf{Model} & MAE & RMSE & MAPE \\
    \midrule
    STOP           & \underline{13.01} & \underline{22.44} & \underline{16.91\%} \\
    STD-PLM        & 14.40 & 25.06 & 18.55\% \\
    \midrule
    \textbf{VaryST}& \textbf{10.49} & \textbf{18.22} & \textbf{14.05\%} \\
    \bottomrule
  \end{tabular}
\end{table}

\begin{table}[!htp]
  \caption{Year-by-year performance on PEMSD3-Stream. 2011: In-Domain Reference}
  \label{tab:pemsd3_pm25_yearly}
  \centering
  \scriptsize
  \setlength{\tabcolsep}{2pt}
  \begin{tabular}{l ccc ccc ccc}
    \toprule
    \multirow{2}{*}{\textbf{Year}} & \multicolumn{3}{c}{\textbf{VaryST}} & \multicolumn{3}{c}{\textbf{STOP}} & \multicolumn{3}{c}{\textbf{STD-PLM}} \\
    \cmidrule(lr){2-4} \cmidrule(lr){5-7} \cmidrule(lr){8-10}
    & MAE & RMSE & MAPE & MAE & RMSE & MAPE & MAE & RMSE & MAPE \\
    \midrule
    2011 & \textbf{10.64} & \textbf{16.47} & \textbf{13.84} & 12.34 & 18.96 & \underline{16.12} & \underline{11.96} & \underline{18.28} & 16.21 \\
    2012 & \textbf{9.97} & \textbf{16.00} & \textbf{13.65} & \underline{11.86} & \underline{18.87} & \underline{16.16} & 13.54 & 22.57 & 17.77 \\
    2013 & \textbf{10.01} & \textbf{16.68} & \textbf{13.60} & \underline{12.27} & \underline{20.05} & \underline{16.26} & 14.24 & 24.16 & 18.23 \\
    2014 & \textbf{10.61} & \textbf{17.39} & \textbf{14.80} & \underline{13.24} & \underline{22.07} & \underline{17.96} & 14.62 & 24.73 & 19.78 \\
    2015 & \textbf{10.55} & \textbf{17.80} & \textbf{14.53} & \underline{12.99} & \underline{22.40} & \underline{17.22} & 14.32 & 24.97 & 18.87 \\
    2016 & \textbf{10.30} & \textbf{21.52} & \textbf{13.23} & \underline{13.19} & \underline{26.20} & \underline{16.14} & 13.89 & 26.77 & 17.46 \\
    2017 & \textbf{11.51} & \textbf{19.93} & \textbf{14.50} & \underline{14.51} & \underline{25.05} & \underline{17.70} & 15.81 & 27.15 & 19.18 \\
    \bottomrule
  \end{tabular}
\end{table}

\subsection{Full Spatial Generalization Results}
\label{app:spatial_ood}
Table~\ref{tab:transfer_zeroshot} presents 12-step zero-shot transfer performance across all twelve pairwise source--target combinations among PEMS03, PEMS04, PEMS07, and PEMS08, evaluated in MAE, RMSE, and percentage MAPE, with the diagonal serving as the in-domain baseline. In parallel, Table~\ref{tab:transfer_fewshot} reports few-shot fine-tuning results using 1\%, 5\%, and 10\% of the target training data. VaryST achieves the lowest MAE, RMSE, and MAPE across all twelve zero-shot transfer directions. Notably, its maximum MAE increase over the corresponding in-domain baseline is confined to just 9.9\% in the PEMS03$\rightarrow$PEMS07 transfer. In fact, transferring from PEMS07 to PEMS03 slightly surpasses the in-domain benchmark, achieving an MAE of 13.78 compared to 13.95. This counter-intuitive advantage aligns with our design: node representations and supports are directly constructed from observations available on the target network rather than rigid source graph structures. Under few-shot adaptation, fine-tuning VaryST on merely 1\% of target data already outperforms STD-PLM adapted with 10\% data across all twelve directions, while scaling the target budget to 10\% yields a modest additional MAE reduction of at most 3.7\%. Figure~\ref{fig:ood_spatial} compares zero-shot and in-domain MAE on PEMS04, with comprehensive analyses provided in Section~\ref{subsec:spatial}.

%
%
\begin{table}[H]
\centering
\caption{Zero-shot cross-dataset transfer (12-step prediction). Rows: source dataset used for full training; columns: target dataset evaluated on. Diagonal cells are the \textcolor{blue}{in-domain (IID)} baselines. Each cell: MAE / RMSE / MAPE (\%).}
\label{tab:transfer_zeroshot}
\resizebox{\textwidth}{!}{%
\footnotesize
\begin{tabular}{l l cccc}
\toprule
\multirow{2}{*}{\textbf{Model}} & \multirow{2}{*}{\textbf{Source}} & \multicolumn{4}{c}{\textbf{Target}} \\
\cmidrule(lr){3-6}
& & \textbf{PEMS03} & \textbf{PEMS04} & \textbf{PEMS07} & \textbf{PEMS08} \\
\midrule
\multirow{4}{*}{\textbf{VaryST}}
& PEMS03 & \textcolor{blue}{13.95 / 22.73 / 13.59} & \underline{18.51 / 30.40 / 12.84} & 20.02 / 32.72 / 8.88  & \underline{13.81 / 22.70 / 9.31} \\
& PEMS04 & \underline{14.03 / 22.77 / 13.44} & \textcolor{blue}{17.56 / 29.35 / 11.81} & \underline{19.91 / 32.64 / 8.48}  & 13.97 / 22.92 / 9.07 \\
& PEMS07 & \textbf{13.78 / 22.32 / 13.24} & \textbf{18.15 / 29.98 / 12.54} & \textcolor{blue}{18.21 / 31.05 / 7.64}  & \textbf{13.46 / 22.44 / 8.86} \\
& PEMS08 & 14.88 / 24.45 / 14.61 & 18.80 / 30.90 / 13.19 & \textbf{19.53 / 32.65 / 8.54}  & \textcolor{blue}{12.89 / 21.99 / 8.50} \\
\midrule
\multirow{4}{*}{STD-PLM}
& PEMS03 & \textcolor{blue}{14.52 / 24.69 / 14.87} & 30.90 / 49.01 / 20.88 & 34.79 / 52.85 / 20.27 & 25.83 / 39.86 / 25.03 \\
& PEMS04 & 23.90 / 38.93 / 27.38 & \textcolor{blue}{18.15 / 30.31 / 11.77} & 35.19 / 53.08 / 23.32 & 26.64 / 40.38 / 28.48 \\
& PEMS07 & 24.98 / 38.37 / 42.77 & 29.44 / 44.95 / 24.05 & \textcolor{blue}{19.26 / 32.72 / 8.09}  & 22.86 / 35.28 / 15.09 \\
& PEMS08 & 22.48 / 36.65 / 36.09 & 28.89 / 44.59 / 22.63 & 31.25 / 48.41 / 14.18 & \textcolor{blue}{13.30 / 23.41 / 8.82}  \\
\bottomrule
\end{tabular}%
}
\end{table}

\begin{table}[H]
\centering
\caption{Few-shot cross-dataset transfer (12-step prediction) with 1\%, 5\% and 10\% of the target training data. Each cell: MAE / RMSE / MAPE (\%).}
\label{tab:transfer_fewshot}
\resizebox{\textwidth}{!}{%
\footnotesize
\begin{tabular}{l l ccc ccc ccc}
\toprule
\multirow{2}{*}{\textbf{Model}} & \multirow{2}{*}{\textbf{Target}}
& \multicolumn{3}{c}{\textbf{1\%}}
& \multicolumn{3}{c}{\textbf{5\%}}
& \multicolumn{3}{c}{\textbf{10\%}} \\
\cmidrule(lr){3-5} \cmidrule(lr){6-8} \cmidrule(lr){9-11}
& & MAE & RMSE & MAPE (\%) & MAE & RMSE & MAPE (\%) & MAE & RMSE & MAPE (\%) \\
\midrule
\multicolumn{11}{l}{\textit{Source: PEMS03}} \\
\multirow{3}{*}{VaryST}  & PEMS04 & \textbf{18.39} & \textbf{30.39} & \textbf{12.60} & \textbf{18.04} & \textbf{29.93} & \textbf{12.33} & \textbf{17.91} & \textbf{29.78} & \textbf{12.19} \\
                        & PEMS07 & \textbf{19.16} & \textbf{32.03} & \textbf{8.16}  & \textbf{18.60} & \textbf{31.48} & \textbf{7.85}  & \textbf{18.63} & \textbf{31.50} & \textbf{7.85}  \\
                        & PEMS08 & \textbf{13.50} & \textbf{22.54} & \textbf{8.91}  & \textbf{13.26} & \textbf{22.31} & \textbf{8.75}  & \textbf{13.17} & \textbf{22.21} & \textbf{8.69}  \\
\cmidrule(lr){2-11}
\multirow{3}{*}{STD-PLM} & PEMS04 & {46.75} & {74.69} & {22.00} & {24.45} & {38.70} & {16.19} & {23.38} & {37.08} & {15.54} \\
                        & PEMS07 & {28.35} & {43.93} & {12.53} & {25.53} & {40.06} & {11.58} & {24.34} & {38.77} & {10.55} \\
                        & PEMS08 & {24.00} & {36.80} & {16.20} & {18.68} & {29.55} & {14.00} & {18.16} & {29.02} & {12.17} \\
\midrule
\multicolumn{11}{l}{\textit{Source: PEMS04}} \\
\multirow{3}{*}{VaryST}  & PEMS03 & \textbf{14.08} & \textbf{22.91} & \textbf{13.49} & \textbf{14.00} & \textbf{22.77} & \textbf{13.35} & \textbf{13.98} & \textbf{22.73} & \textbf{13.45} \\
                        & PEMS07 & \textbf{19.27} & \textbf{32.14} & \textbf{8.11}  & \textbf{18.74} & \textbf{31.62} & \textbf{7.86}  & \textbf{18.65} & \textbf{31.49} & \textbf{7.82}  \\
                        & PEMS08 & \textbf{13.64} & \textbf{22.69} & \textbf{8.77}  & \textbf{13.34} & \textbf{22.36} & \textbf{8.65}  & \textbf{13.20} & \textbf{22.21} & \textbf{8.59}  \\
\cmidrule(lr){2-11}
\multirow{3}{*}{STD-PLM} & PEMS03 & {20.30} & {32.79} & {21.97} & {18.35} & {30.37} & {18.00} & {17.01} & {27.84} & {16.67} \\
                        & PEMS07 & {27.60} & {42.84} & {12.58} & {25.17} & {39.70} & {11.35} & {24.55} & {39.07} & {10.72} \\
                        & PEMS08 & {21.99} & {33.25} & {15.84} & {18.47} & {29.40} & {12.70} & {18.05} & {28.96} & {11.80} \\
\midrule
\multicolumn{11}{l}{\textit{Source: PEMS07}} \\
\multirow{3}{*}{VaryST}  & PEMS03 & \textbf{13.91} & \textbf{22.56} & \textbf{13.31} & \textbf{13.89} & \textbf{22.54} & \textbf{13.24} & \textbf{13.88} & \textbf{22.55} & \textbf{13.18} \\
                        & PEMS04 & \textbf{18.20} & \textbf{30.12} & \textbf{12.50} & \textbf{17.88} & \textbf{29.75} & \textbf{12.15} & \textbf{17.87} & \textbf{29.72} & \textbf{12.18} \\
                        & PEMS08 & \textbf{13.27} & \textbf{22.31} & \textbf{8.77}  & \textbf{13.09} & \textbf{22.18} & \textbf{8.60}  & \textbf{13.02} & \textbf{22.09} & \textbf{8.62}  \\
\cmidrule(lr){2-11}
\multirow{3}{*}{STD-PLM} & PEMS03 & {20.41} & {33.44} & {26.46} & {18.35} & {30.69} & {18.23} & {16.86} & {27.78} & {16.38} \\
                        & PEMS04 & {32.67} & {53.73} & {20.74} & {24.07} & {38.02} & {16.76} & {23.19} & {36.79} & {15.88} \\
                        & PEMS08 & {21.78} & {33.36} & {18.84} & {18.18} & {29.20} & {11.53} & {18.08} & {29.17} & {11.37} \\
\midrule
\multicolumn{11}{l}{\textit{Source: PEMS08}} \\
\multirow{3}{*}{VaryST}  & PEMS03 & \textbf{14.78} & \textbf{24.57} & \textbf{14.41} & \textbf{14.56} & \textbf{24.10} & \textbf{14.21} & \textbf{14.57} & \textbf{23.95} & \textbf{14.16} \\
                        & PEMS04 & \textbf{18.66} & \textbf{30.98} & \textbf{12.84} & \textbf{18.12} & \textbf{30.07} & \textbf{12.46} & \textbf{17.97} & \textbf{29.87} & \textbf{12.27} \\
                        & PEMS07 & \textbf{19.01} & \textbf{32.15} & \textbf{8.08}  & \textbf{18.70} & \textbf{31.83} & \textbf{7.89}  & \textbf{18.51} & \textbf{31.53} & \textbf{7.81}  \\
\cmidrule(lr){2-11}
\multirow{3}{*}{STD-PLM} & PEMS03 & {19.69} & {32.93} & {23.70} & {18.38} & {31.25} & {17.96} & {16.91} & {28.28} & {16.31} \\
                        & PEMS04 & {47.90} & {84.18} & {23.60} & {24.44} & {38.50} & {16.66} & {23.57} & {37.27} & {15.98} \\
                        & PEMS07 & {27.63} & {42.92} & {12.31} & {25.38} & {39.94} & {11.14} & {24.38} & {38.94} & {10.59} \\
\bottomrule
\end{tabular}%
}
\end{table}


\subsection{Full Missing-Observation Results}
\label{app:miss_ood}

The evaluation covers forecasting on PEMS04 and PEMS08 under point missing (RM) and spatio-temporal block missing (CM) at missing rates of 10\% to 70\%, with all metrics averaged over the 12 horizons at observed positions. VaryST trains one model per rate with importance-guided observation perturbation (IGOP), which removes the history patches that draw the most attention during training, whereas baseline inputs are pre-filled by linear interpolation, following Appendix~\ref{app:missing_protocol}. Table~\ref{tab:missing} reports same-rate training and testing.
\begin{table}[H]
\centering
\caption{Overall prediction performance with missing observations on PEMS04 and PEMS08 under missing rates from 10\% to 70\%. The best and second-best results are in \textbf{bold} and \underline{underline}, respectively.}
\label{tab:missing}
\setlength{\tabcolsep}{3pt}
\resizebox{\textwidth}{!}{%
\footnotesize
\begin{tabular}{l ccc ccc ccc ccc}
\toprule
\multirow{2}{*}{\textbf{Model}}
& \multicolumn{3}{c}{\textbf{10\%}} & \multicolumn{3}{c}{\textbf{30\%}} & \multicolumn{3}{c}{\textbf{50\%}} & \multicolumn{3}{c}{\textbf{70\%}} \\
\cmidrule(lr){2-4} \cmidrule(lr){5-7} \cmidrule(lr){8-10} \cmidrule(lr){11-13}
& MAE & RMSE & MAPE (\%) & MAE & RMSE & MAPE (\%) & MAE & RMSE & MAPE (\%) & MAE & RMSE & MAPE (\%) \\
\midrule
\multicolumn{13}{c}{\textbf{PEMS04 Point Missing (RM)}} \\
\midrule
FaST$^\dagger$ & 18.67 & 30.29 & 12.89 & 18.74 & 30.37 & 13.01 & 18.87 & 30.50 & 13.05 & 19.17 & 30.78 & \underline{13.14} \\
STID$^\dagger$ & \underline{18.34} & \underline{29.92} & 13.09 & \underline{18.38} & 29.98 & \underline{12.84} & \underline{18.53} & 30.20 & \underline{12.99} & \underline{18.75} & \underline{30.40} & 13.41 \\
BiST$^\dagger$ & 18.47 & 29.94 & \underline{12.88} & 18.53 & \underline{29.94} & 13.25 & \underline{18.53} & \underline{30.10} & 13.22 & 19.00 & 30.42 & 13.69 \\
\textbf{VaryST}  & \textbf{17.60} & \textbf{29.34} & \textbf{11.90} & \textbf{17.40} & \textbf{29.20} & \textbf{11.74} & \textbf{17.47} & \textbf{29.28} & \textbf{11.80} & \textbf{17.70} & \textbf{29.39} & \textbf{11.93} \\
\midrule
\multicolumn{13}{c}{\textbf{PEMS04 Spatio-Temporal Block Missing (CM)}} \\
\midrule
FaST$^\dagger$ & 18.79 & 30.32 & \underline{13.07} & 20.39 & 32.74 & 14.16 & 22.02 & 34.70 & 15.99 & 27.17 & 41.55 & 20.67 \\
STID$^\dagger$ & 18.59 & 30.22 & 13.48 & 19.66 & 32.20 & 13.84 & 21.08 & 33.92 & 14.88 & 23.54 & 38.39 & \underline{16.08} \\
BiST$^\dagger$ & \underline{18.58} & \underline{30.13} & 13.51 & \underline{19.40} & \underline{31.39} & \underline{13.80} & \underline{20.70} & \underline{33.06} & \underline{14.87} & \underline{23.09} & \underline{36.81} & 17.41 \\
\textbf{VaryST}  & \textbf{17.80} & \textbf{29.55} & \textbf{12.05} & \textbf{18.44} & \textbf{30.42} & \textbf{12.48} & \textbf{19.50} & \textbf{31.76} & \textbf{13.39} & \textbf{22.41} & \textbf{36.08} & \textbf{15.29} \\
\midrule
\multicolumn{13}{c}{\textbf{PEMS08 Point Missing (RM)}} \\
\midrule
FaST$^\dagger$ & 14.15 & 23.26 & 10.08 & 14.18 & 23.36 & 10.14 & 14.29 & 23.65 & 10.18 & 14.77 & 23.97 & 10.47 \\
STID$^\dagger$ & \underline{13.50} & \underline{22.50} & \underline{9.34} & \underline{13.61} & \underline{22.53} & \underline{9.73} & \underline{13.73} & \underline{22.62} & \underline{9.60} & \underline{14.08} & \underline{23.12} & \underline{10.12} \\
BiST$^\dagger$ & 13.99 & 22.74 & 10.75 & 13.92 & 22.77 & 10.59 & 14.02 & 22.94 & 10.33 & 14.27 & 23.24 & 10.53 \\
\textbf{VaryST}  & \textbf{12.98} & \textbf{22.13} & \textbf{8.61} & \textbf{12.91} & \textbf{21.91} & \textbf{8.56} & \textbf{12.96} & \textbf{21.96} & \textbf{8.55} & \textbf{13.30} & \textbf{22.27} & \textbf{8.73} \\
\midrule
\multicolumn{13}{c}{\textbf{PEMS08 Spatio-Temporal Block Missing (CM)}} \\
\midrule
FaST$^\dagger$ & 14.42 & 23.75 & 10.40 & 15.40 & 25.00 & 11.00 & 17.48 & 27.69 & 12.93 & 22.08 & 33.78 & 17.09 \\
STID$^\dagger$ & \underline{13.74} & \underline{22.80} & \underline{9.60} & \underline{14.73} & \underline{23.83} & \underline{10.35} & \underline{16.12} & \underline{25.91} & 12.12 & \underline{17.96} & \underline{28.94} & \underline{12.71} \\
BiST$^\dagger$ & 14.06 & 23.12 & 10.54 & 15.03 & 24.14 & 10.83 & 16.31 & 26.19 & \underline{11.80} & 18.51 & 29.11 & 14.22 \\
\textbf{VaryST}  & \textbf{13.14} & \textbf{22.32} & \textbf{8.63} & \textbf{14.30} & \textbf{23.38} & \textbf{9.63} & \textbf{14.89} & \textbf{24.67} & \textbf{9.72} & \textbf{17.16} & \textbf{28.08} & \textbf{11.56} \\
\bottomrule
\end{tabular}%
}
\vspace{2pt}
\begin{minipage}{\textwidth}
\footnotesize
\raggedright
$^\dagger$ Models adapted to long-context inputs (2016 steps) by us for fair comparison. \\
\end{minipage}
\end{table}

Under this protocol VaryST attains the lowest metric in all 48 reported cells. Scattered point missing barely affects it: between 10\% and 70\% RM the same-rate MAE stays within 0.4 on both datasets, because the encoder consumes the availability mask directly and re-derives node representations from the observations that remain. Block missing is the harder regime: as entire segments vanish, the error of every model grows with the rate, and VaryST remains first throughout.

\begin{table}[H]
\centering
\caption{Generalization to unseen missing rates on PEMS04 and PEMS08. Models are trained at a 30\% missing rate.}
\label{tab:missing_cross_rate}
\setlength{\tabcolsep}{3pt}
\resizebox{\textwidth}{!}{%
\footnotesize
\begin{tabular}{l ccc ccc ccc ccc}
\toprule
\multirow{2}{*}{\textbf{Model}}
& \multicolumn{3}{c}{\textbf{10\%}} & \multicolumn{3}{c}{\textbf{30\%}} & \multicolumn{3}{c}{\textbf{50\%}} & \multicolumn{3}{c}{\textbf{70\%}} \\
\cmidrule(lr){2-4} \cmidrule(lr){5-7} \cmidrule(lr){8-10} \cmidrule(lr){11-13}
& MAE & RMSE & MAPE (\%) & MAE & RMSE & MAPE (\%) & MAE & RMSE & MAPE (\%) & MAE & RMSE & MAPE (\%) \\
\midrule
\multicolumn{13}{c}{\textbf{PEMS04 Point Missing (RM), source = 30\%}} \\
\midrule
FaST$^\dagger$ & 18.69 & 30.31 & 12.97 & 18.74 & 30.37 & 13.01 & 18.88 & 30.59 & 13.09 & 19.22 & 30.97 & 13.39 \\
STID$^\dagger$ & \underline{18.32} & 29.91 & \underline{12.80} & \underline{18.38} & 29.98 & \underline{12.84} & \underline{18.53} & 30.19 & \underline{12.91} & \underline{18.91} & \underline{30.62} & \underline{13.19} \\
BiST$^\dagger$ & 18.49 & \underline{29.90} & 13.22 & 18.53 & \underline{29.94} & 13.25 & 18.66 & \underline{30.12} & 13.32 & 18.98 & \textbf{30.12} & 13.57 \\
\textbf{VaryST}  & \textbf{17.60} & \textbf{29.35} & \textbf{12.07} & \textbf{17.40} & \textbf{29.20} & \textbf{11.74} & \textbf{17.55} & \textbf{29.38} & \textbf{11.73} & \textbf{18.28} & \textbf{30.12} & \textbf{12.34} \\
\midrule
\multicolumn{13}{c}{\textbf{PEMS04 Spatio-Temporal Block Missing (CM), source = 30\%}} \\
\midrule
FaST$^\dagger$ & 19.88 & 32.16 & 13.63 & 20.39 & 32.74 & 14.16 & 23.37 & 36.43 & 18.95 & 38.20 & 57.56 & 39.79 \\
STID$^\dagger$ & 19.07 & 31.42 & \underline{13.21} & 19.66 & 32.20 & 13.84 & 22.27 & 35.67 & 17.73 & \underline{34.73} & \underline{52.86} & \underline{35.44} \\
BiST$^\dagger$ & \underline{18.98} & \underline{30.86} & 13.42 & \underline{19.40} & \underline{31.39} & \underline{13.80} & \underline{21.73} & \underline{34.31} & \underline{17.31} & 34.83 & \textbf{52.23} & 36.97 \\
\textbf{VaryST}  & \textbf{18.02} & \textbf{29.91} & \textbf{12.35} & \textbf{18.44} & \textbf{30.42} & \textbf{12.48} & \textbf{20.61} & \textbf{34.15} & \textbf{14.31} & \textbf{34.71} & 58.39 & \textbf{32.20} \\
\midrule
\multicolumn{13}{c}{\textbf{PEMS08 Point Missing (RM), source = 30\%}} \\
\midrule
FaST$^\dagger$ & 14.11 & 23.28 & 10.09 & 14.18 & 23.36 & 10.14 & 14.29 & 23.50 & 10.22 & 14.68 & 24.00 & 10.43 \\
STID$^\dagger$ & \underline{13.55} & \underline{22.47} & \underline{9.70} & \underline{13.61} & \underline{22.53} & \underline{9.73} & \underline{13.76} & \underline{22.71} & \underline{9.82} & \underline{14.18} & \underline{23.26} & \underline{10.02} \\
BiST$^\dagger$ & 13.88 & 22.74 & 10.58 & 13.92 & 22.77 & 10.59 & 14.03 & 22.90 & 10.68 & 14.42 & 23.44 & 10.88 \\
\textbf{VaryST}  & \textbf{13.05} & \textbf{22.05} & \textbf{8.73} & \textbf{12.91} & \textbf{21.91} & \textbf{8.56} & \textbf{13.07} & \textbf{22.05} & \textbf{8.62} & \textbf{13.83} & \textbf{22.87} & \textbf{9.02} \\
\midrule
\multicolumn{13}{c}{\textbf{PEMS08 Spatio-Temporal Block Missing (CM), source = 30\%}} \\
\midrule
FaST$^\dagger$ & 14.73 & 24.16 & 10.47 & 15.40 & 25.00 & 11.00 & 18.57 & 29.78 & 13.30 & 32.06 & 48.39 & 26.39 \\
STID$^\dagger$ & \underline{14.05} & \underline{23.00} & \underline{9.82} & \underline{14.73} & \underline{23.83} & \underline{10.35} & 17.31 & 28.23 & \underline{12.18} & 28.72 & \underline{46.28} & 21.98 \\
BiST$^\dagger$ & 14.44 & 23.40 & 10.34 & 15.03 & 24.14 & 10.83 & \underline{17.30} & \underline{27.90} & 12.53 & \textbf{27.43} & \textbf{43.88} & \textbf{21.04} \\
\textbf{VaryST}  & \textbf{13.84} & \textbf{22.83} & \textbf{9.21} & \textbf{14.30} & \textbf{23.38} & \textbf{9.63} & \textbf{16.42} & \textbf{27.82} & \textbf{11.18} & \underline{27.76} & 48.80 & \underline{21.84} \\
\bottomrule
\end{tabular}%
}
\vspace{2pt}
\begin{minipage}{\textwidth}
\footnotesize
\raggedright
$^\dagger$ Models adapted to long-context inputs (2016 steps) by us for fair comparison.
\end{minipage}
\end{table}
Table~\ref{tab:missing_cross_rate} reports cross-rate transfer performance, where models trained at a 30\% missing rate are directly evaluated across rates from 10\% to 70\% without retraining or altering the missing pattern. Under RM, VaryST ranks first across all 24 evaluation cells. Even at the extreme 70\% missing rate, its MAE remains 0.63 and 0.35 lower than the runner-up on PEMS04 and PEMS08, respectively. Although its degradation relative to same-rate training reaches +0.58 on PEMS04 and +0.53 on PEMS08, surpassing the baselines' maximum gap of 0.16, scattered observations still suffice to reconstruct its representations. Under CM, VaryST maintains its top ranking up to a 50\% missing rate; at 70\%, where contiguous evidence largely vanishes, it still achieves the best MAE and MAPE on PEMS04, outperforming the runner-up by 3.2 points in MAPE. On PEMS08 it is second to BiST on MAE, 27.76 against 27.43, and it is last on RMSE on both datasets.

\begin{table}[H]
\centering
\caption{Cross-pattern generalization at 30\% missing rate on PEMS04 and PEMS08.}
\label{tab:missing_cross_pattern}
\setlength{\tabcolsep}{3pt}
\resizebox{\textwidth}{!}{%
\footnotesize
\begin{tabular}{l ccc ccc ccc ccc}
\toprule
\multirow{2}{*}{\textbf{Model}}
& \multicolumn{6}{c}{\textbf{PEMS04}} & \multicolumn{6}{c}{\textbf{PEMS08}} \\
\cmidrule(lr){2-7} \cmidrule(lr){8-13}
& \multicolumn{3}{c}{RM $\rightarrow$ CM} & \multicolumn{3}{c}{CM $\rightarrow$ RM} & \multicolumn{3}{c}{RM $\rightarrow$ CM} & \multicolumn{3}{c}{CM $\rightarrow$ RM} \\
\cmidrule(lr){2-4} \cmidrule(lr){5-7} \cmidrule(lr){8-10} \cmidrule(lr){11-13}
& MAE & RMSE & MAPE (\%) & MAE & RMSE & MAPE (\%) & MAE & RMSE & MAPE (\%) & MAE & RMSE & MAPE (\%) \\
\midrule
FaST$^\dagger$ & 20.32 & 32.40 & 15.19 & 19.89 & 32.16 & 13.62 & 16.00 & 26.00 & 11.87 & 14.71 & 24.13 & 10.46 \\
STID$^\dagger$ & \underline{19.96} & 32.14 & \underline{14.71} & 19.06 & 31.41 & \underline{13.13} & \underline{15.33} & \textbf{24.85} & \underline{11.28} & \textbf{14.02} & \textbf{22.95} & \underline{9.80} \\
BiST$^\dagger$ & 20.02 & \textbf{31.87} & 15.06 & \underline{18.98} & \underline{30.86} & 13.38 & 15.63 & 25.06 & 12.33 & 14.42 & 23.37 & 10.31 \\
\textbf{VaryST}  & \textbf{19.51} & \underline{31.94} & \textbf{13.96} & \textbf{18.40} & \textbf{30.23} & \textbf{12.75} & \textbf{15.00} & \underline{24.93} & \textbf{10.43} & \underline{14.14} & \underline{23.15} & \textbf{9.36} \\
\midrule
\multicolumn{13}{l}{\emph{In-domain reference at 30\% (train and test on the target pattern)}} \\
STID$^\dagger$ & 19.66 & 32.20 & 13.84 & 18.38 & 29.98 & 12.84 & 14.73 & 23.83 & 10.35 & 13.61 & 22.53 & 9.73 \\
BiST$^\dagger$ & 19.40 & 31.39 & 13.80 & 18.53 & 29.94 & 13.25 & 15.03 & 24.14 & 10.83 & 13.92 & 22.77 & 10.59 \\
\bottomrule
\end{tabular}%
}
\vspace{2pt}
\begin{minipage}{\textwidth}
\footnotesize
\raggedright
$^\dagger$ Models adapted to long-context inputs (2016 steps) by us for fair comparison.
\end{minipage}
\end{table}
Table~\ref{tab:missing_cross_pattern} reports cross-pattern transfer at a 30\% missing rate: a model trained under one pattern is evaluated under the other without further training, and the reference rows repeat the in-domain numbers at 30\% from Table~\ref{tab:missing}. VaryST attains the lowest MAE in three of the four directions and the lowest MAPE in all four, as its representations are re-derived from the observations that survive the switch. Its RMSE is second in both RM$\rightarrow$CM directions by at most 0.08, and STID leads MAE and RMSE in the CM$\rightarrow$RM direction on PEMS08 (14.14 vs.\ 14.02 and 23.15 vs.\ 22.95), consistent with STID modeling each node independently and being least sensitive to the pattern switch.

The rate and pattern sweeps are summarized in Figure~\ref{fig:ood_missing} of the main text.

\subsection{Full Ablation Results}
\label{app:full_ablation}

The complete ablation in Table~\ref{tab:ablation} covers 12-step prediction on PEMS03 and PEMS08, reporting MAE, RMSE, MAPE, and MAPE$_{10}$, where MAPE$_{10}$ excludes ground-truth values below 10. Removing any component raises the MAE and RMSE on PEMS03. The raw MAPE deserves care. Readings below 10 carry little operational value in traffic applications, but their small denominators make relative errors fluctuate, so a few variants look marginally better under this metric. Once these low-value readings are excluded, the full model attains the best MAPE$_{10}$ on both datasets, and their apparent advantages prove to be small-value artifacts rather than genuine improvements. The hop-aware structural attention causes the largest PEMS03 degradation at 0.47 MAE, and removing the global token costs 0.33 MAE on PEMS08, identifying structural modulation and global-state completion as the main sources of in-domain accuracy. The PEMS03 breakdown appears in Figure~\ref{fig:ablation_pems03} of the main text.


\begin{table}[H]
\centering
\caption{Ablation study of each component on PEMS03 and PEMS08 (12-step prediction).}
\label{tab:ablation}
\setlength{\tabcolsep}{5pt}
\footnotesize
\begin{tabular}{l cccc cccc}
\toprule
\multirow{2}{*}{\textbf{Variant}}
& \multicolumn{4}{c}{\textbf{PEMS03}} & \multicolumn{4}{c}{\textbf{PEMS08}} \\
\cmidrule(lr){2-5} \cmidrule(lr){6-9}
& MAE & RMSE & MAPE & MAPE$_{10}$ & MAE & RMSE & MAPE & MAPE$_{10}$ \\
\midrule
VaryST (full)            & \textbf{13.95} & \textbf{22.73} & 13.59 & \textbf{11.00} & \underline{12.89} & \underline{21.99} & \underline{8.50} & \textbf{7.29} \\
\midrule
w/o Laplacian PE         & 14.13 & 23.02 & \underline{13.47} & \underline{11.04} & 12.91 & 22.00 & 8.57 & 7.31 \\
w/o Hop-RoPE             & 14.42 & 23.75 & 13.93 & 11.40 & 12.93 & 22.07 & 8.56 & 7.32 \\
w/o structural encoding  & 14.10 & 23.05 & 13.55 & 11.11 & 12.92 & 22.06 & 8.54 & \underline{7.30} \\
w/o recent-12 token      & 14.10 & 23.00 & \textbf{13.35} & \textbf{10.99} & \textbf{12.86} & \textbf{21.97} & \textbf{8.47} & \textbf{7.29} \\
w/o long context          & 15.95 & 25.42 & 15.73 & 13.04 & 15.02 & 25.51 & 9.81 & 8.45 \\
w/o global token         & \underline{14.06} & \underline{22.97} & \underline{13.47} & \underline{11.04} & 13.22 & 22.23 & 8.67 & 7.47 \\
Cosine retrieval           & 14.33 & 23.31 & 13.84 & 11.32 & \underline{12.88} & 22.02 & \underline{8.49} & \underline{7.30} \\
w/o tod/dow embedding    & 14.20 & 23.04 & 13.74 & 11.22 & 12.98 & 22.07 & 8.62 & 7.35 \\
\bottomrule
\end{tabular}
\end{table}


\subsection{Full Hyperparameter Study Results}\label{app:hyperparam_full}

Table~\ref{tab:hyperparam} presents the complete hyperparameter study. We evaluate 12-step prediction on PEMS03 and PEMS08 by sweeping each hyperparameter against the default configuration in Table~\ref{tab:default_hyperparam}: context length $T$ across 36, 288, 864, 2016, and 4032 steps; patch length across 12, 36, and 96 steps; and the number of retrieved support nodes $K$ across 16, 32, 64, and 128. Figure~\ref{fig:hyperparam_pems03} illustrates the corresponding curves for PEMS03.

Context length exerts the strongest influence, characterized by a sharp initial error reduction that rapidly plateaus. Increasing $T$ from 36 to 288 steps substantially lowers MAE by 1.04 on PEMS03 and 1.18 on PEMS08. Beyond 288 steps, the error declines gradually toward the one-week default ($T = 2016$), reaching MAEs of 13.95 and 12.89, respectively. Extending the context to two weeks ($T = 4032$) yields no further gain on PEMS08 and degrades MAE by 0.45 on PEMS03, confirming the default as the optimal operating point. This behavior directly reflects our representation design: compressing a 2016-step history into a single observation-derived token per node enriches the evidence underlying each token without expanding the $(K{+}3)$-token decoder sequence, up to the point of diminishing informational returns.

Spatial retrieval exhibits a similar economy. Even with only 16 support nodes, the MAE remains within 0.42 of the optimum on both benchmarks, whereas scaling $K$ to 128 brings no consistent benefit. The RBF retrieval effectively concentrates relevant spatial signals into a small subset of nodes, while formatting them as prompt tokens preserves linear decoding complexity in $K$ rather than quadratic scaling in total node count. Finally, patch length is the least sensitive hyperparameter; varying it across 12, 36, and 96 steps shifts MAE by at most 0.35 on both datasets, demonstrating that the representation is largely agnostic to the exact historical segmentation.

\begin{table}[H]
\centering
\caption{Hyperparameter study on PEMS03 and PEMS08. The best and second-best results are in \textbf{bold} and \underline{underline}, respectively.}
\label{tab:hyperparam}
\setlength{\tabcolsep}{5pt}
\footnotesize
\begin{tabular}{l ccc ccc}
\toprule
\multirow{2}{*}{\textbf{Setting}}
& \multicolumn{3}{c}{\textbf{PEMS03}} & \multicolumn{3}{c}{\textbf{PEMS08}} \\
\cmidrule(lr){2-4} \cmidrule(lr){5-7}
& MAE & RMSE & MAPE & MAE & RMSE & MAPE \\
\midrule
\multicolumn{7}{l}{\textit{Input context length}} \\
\quad 36               & 15.30 & 24.37 & 14.88 & 14.78 & 24.91 & 9.67 \\
\quad 288              & \underline{14.26} & \underline{23.08} & 13.87 & 13.60 & 22.79 & 9.02 \\
\quad 864              & 14.31 & 23.37 & \underline{13.65} & 13.40 & 22.53 & 8.90 \\
\quad \textbf{2016 (default)} & \textbf{13.95} & \textbf{22.73} & \textbf{13.59} & \textbf{12.89} & \underline{21.99} & \textbf{8.50} \\
\quad 4032             & 14.40 & 23.74 & 13.87 & \underline{12.89} & \textbf{21.97} & \underline{8.51} \\
\midrule
\multicolumn{7}{l}{\textit{Patch length}} \\
\quad 12               & \underline{14.03} & \underline{22.90} & \textbf{13.47} & 13.23 & 22.25 & 8.80 \\
\quad \textbf{36 (default)}   & \textbf{13.95} & \textbf{22.73} & \underline{13.59} & \textbf{12.89} & \textbf{21.99} & \textbf{8.50} \\
\quad 96               & 14.26 & 23.25 & 13.95 & \underline{13.13} & \underline{22.24} & \underline{8.75} \\
\midrule
\multicolumn{7}{l}{\textit{Number of auxiliary nodes}} \\
\quad 16               & 14.37 & 23.33 & 13.94 & 12.89 & 22.06 & \underline{8.49} \\
\quad 32               & \underline{14.11} & \underline{23.02} & \underline{13.64} & 12.93 & 22.02 & 8.54 \\
\quad \textbf{64 (default)}   & \textbf{13.95} & \textbf{22.73} & \textbf{13.59} & \underline{12.89} & \underline{21.99} & 8.50 \\
\quad 128              & 14.17 & 23.14 & 13.72 & \textbf{12.88} & \textbf{21.92} & \textbf{8.48} \\
\bottomrule
\end{tabular}
\end{table}

\end{document}